\documentclass{article}

    \PassOptionsToPackage{numbers, compress}{natbib}

\usepackage[final]{neurips_2025}

\usepackage[utf8]{inputenc} 
\usepackage[T1]{fontenc}    
\usepackage[colorlinks=true, urlcolor=blue]{hyperref}       
\usepackage{url}            
\usepackage{booktabs}       
\usepackage{amsfonts}       
\usepackage{nicefrac}       
\usepackage{microtype}      
\usepackage{xcolor}         
\usepackage{graphicx}
\usepackage{bbm}
\usepackage{subcaption}
\usepackage{amsmath}
\usepackage{bm}
\usepackage{multicol}
\usepackage{multirow}
\usepackage{graphicx}
\usepackage{wrapfig}
\usepackage{float}
\usepackage{verbatim}

\usepackage{natbib}

\title{Exploiting Vocabulary Frequency Imbalance \\ in Language Model Pre-training}

\author{%
  Woojin Chung\thanks{Equal contribution} \\ 
  KAIST\thanks{Korea Advanced Institute of Science and Technology (KAIST)} \\ 
  \texttt{gartland223@gmail.com} \\
  \And
  Jeonghoon Kim\footnotemark[1] \ \thanks{Corresponding author} \\
  NAVER Cloud \& KAIST \\
  \texttt{jeonghoon.samuel@gmail.com} \\
}

\begin{document}

\maketitle

\newcommand{\iconlink}[3]{%
  \raisebox{-0.2ex}{\includegraphics[height=1.1em]{#1}}\,%
  \href{#2}{#3}%
}

\begin{center}
\small

\iconlink{styles/figure/gitHub.png}
         {https://github.com/Chung-Kim/vocab-imbalance}
         {github.com/Chung-Kim/vocab-imbalance}\\[0.4em]
\iconlink{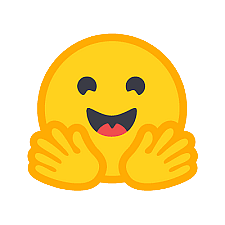}
         {https://huggingface.co/collections/gartland/neurips-2025-vocabulary-frequency-imbalance}
         {huggingface.co/collections/gartland/neurips-2025-vocabulary-frequency-imbalance}
\end{center}

\begin{abstract}

Large language models are trained with tokenizers, and the resulting token distribution is highly \emph{imbalanced}: a few words dominate the stream while most occur rarely. Recent practice favors ever-larger vocabularies, but it is unclear where the benefit comes from. To this end, we perform a controlled study that scales the vocabulary of the language model from \mbox{$24$K} to \mbox{$196$K} while holding data, computation, and optimization unchanged. We begin by quantifying the complexity of tokenized text -- formalized via Kolmogorov complexity -- and show that larger vocabularies reduce this complexity. Above \mbox{$24$K}, every common word is already tokenized as a single token, so enlarging vocabulary only deepens the relative token-frequency imbalance. Word-level loss decomposition shows that larger vocabularies reduce cross-entropy loss almost exclusively by lowering uncertainty on the $2,500$ most frequent words, even though loss on the rare tail rises. Same frequent words cover roughly $75\%$ of tokens in downstream benchmarks, this training advantage transfers intact. We further show that enlarging model parameters with a fixed vocabulary yields the same frequent-word benefit. Our results recast “bigger vocabularies help” as “lowering complexity of tokenized text helps,” offering a simple, principled knob for tokenizer–model co-design and clarifying the loss dynamics that govern language model scaling in pre-training. 

\end{abstract}

\section{Introduction}

A language model incorporates a tokenizer that converts a stream of characters into a series of token IDs, each representing a specific substring \citep{huang2025overtokenizedtransformervocabularygenerally,rajaraman2025theorytokenizationllms,dagan2024gettingtokenizerpretrainingdomain}. Tokenization has re-emerged as an powerful tuning knob for language models, with mounting evidence that simply scaling up vocabulary consistently reduces perplexity and improves downstream accuracy across diverse domains and model scales \citep{tao2024scalinglawsvocabularylarger,yu2025}. Although this trend is consistently observed in practice, the underlying mechanism responsible for it has yet to be thoroughly investigated.

As vocabulary size grows, adding merge candidates segments frequent words in the training data into a single token and pushes infrequent ones deeper into the long tail, sharpening the relative token‑frequency distribution \citep{schmidt2025boundlessbytepairencoding,superbpe,hayase2024data}. In unigram models, this simply lowers the unigram Shannon entropy of training data toward its optimal loss (i.e., entropy rate) \cite{rajaraman2025theorytokenizationllms}. This intuition does not transfer to language models as they condition their next-token prediction on the preceding context, which contains a mixture of common and rare tokens \cite{benjio2000,gpt-3,gpt-4,gpt-neox}.  Moreover, rare tokens already carry much lower conditional probabilities than marginal ones; mistakes on them incur disproportionately large penalties \cite{pinto2024fairlanguagemodelparadox,mircea2024gradient}. Yet rare tokens account for a small share of the entire dataset, making it unclear how a larger vocabulary reallocates capacity between frequent and rare tokens.

In this study, we explore why enlarging vocabulary size improves the performance of language models by expanding vocabulary from $24$K to $196$K. Viewing BPE \cite{BPE} with its pre-tokenization rules as a lossless compressor \cite{deletang_modeling,Lester2024}, we assess how much the tokenized text is compressed against raw text by an upper-bound on Kolmogorov complexity and first illustrate that expanding the vocabulary reduces this complexity (\S\ref{main:complexity}). Above $24$K, frequent words are already encoded as single tokens, so the primary shift is a heightened imbalance in relative token-frequency distribution (\S\ref{main:segment efficiency}), regardless of dataset quality (\S\ref{main:dataset quality}). Furthermore, a word-level loss decomposition shows that a larger vocabulary reduces the loss of frequent words, thereby lowering the model’s overall cross-entropy (\S\ref{main:token freq}), regardless of dataset and model size. 

Analytic experiments trace how enlarging the vocabulary changes both training dynamics and generalization behavior through token-frequency imbalance. Our observation suggests that high frequency words in the pre-training corpus largely coincide with those in downstream benchmarks, both in identity and coverage, explaining the close link between training loss and transfer accuracy (\S\ref{main:generalization}). Scaling the model itself produces the same benefit: it predicts frequent words more accurately, thereby enhancing overall language‑model performance (\S\ref{main:model size}). 

\textbf{Contribution.} We identify that larger vocabularies reduce the complexity of tokenized text, thereby facilitating the model to learn non-i.i.d. patterns in the training data and lowering language modeling difficulty. Our experiments further reveal that beyond a certain size, vocabulary expansion no longer improves segmentation but instead steepens the skewness of the token‐frequency distribution. This sharper imbalance alone lowers global cross‐entropy by reducing the top $\sim2{,}500$ frequent words loss despite slight degradation on the rare tail. Through cross‐dataset overlap analyses, we demonstrate that exploiting, rather than mitigating, token frequency imbalance causally reduces cross‐entropy and boosts downstream accuracy. Finally, we show that parameter scaling replicates the same benefit as vocabulary scaling, both primarily reduce uncertainty on the same set of frequent tokens.


\section{Motivation}\label{motivations}
Tokenization -- the interface between raw text and the discrete symbols a model actually sees -- has resurfaced as a powerful, low-cost lever for improving language model quality \citep{huang2025overtokenizedtransformervocabularygenerally,rajaraman2025theorytokenizationllms,dagan2024gettingtokenizerpretrainingdomain,hayase2024data}. A growing body of evidence finds that simply \textit{increasing the size of the tokenizer vocabulary} yields systematic gains in perplexity and downstream accuracy across domain and model scales \citep{tao2024scalinglawsvocabularylarger,yu2025}. Despite this empirical regularity, the mechanism behind the gain remains underexplored.  

The clue may lie in how tokenizers behave as their vocabularies grow. \citet{rajaraman2025theorytokenizationllms} noted that adding merge candidates segments the most frequent words in the corpus into single tokens, making an i.i.d. model over tokens a closer approximation to inherently non-i.i.d. data. Moreover, they empirically confirm that larger vocabularies lower unigram cross-entropy. Simultaneously, this pushes rare vocabularies further into the long tail, yielding a markedly more skewed token-frequency distribution (after the usual whitespace pre-tokenization) \citep{schmidt2025boundlessbytepairencoding,superbpe}. 


Unfortunately, this explanation does not carry over verbatim to neural language models. In language models, every prediction is conditioned on a variable-length context. Errors on rare vocabularies are disproportionately costly because their conditional probabilities are already orders of magnitude below their marginal frequencies \cite{pinto2024fairlanguagemodelparadox,mircea2024gradient}, yet they occupy only a tiny fraction of the overall loss. Quantifying how vocabulary growth redistributes modelling capacity between these two regimes, therefore, remains a non-trivial challenge (see the Appendix \ref{apdx:motivation} for a detailed explanation).

In this work, we tackle the problem head-on by pairing large-scale controlled experiments with analytical diagnostics. We ask:

\begin{quote}
\centering
\emph{Why does enlarging the tokenizer vocabulary improve Transformer performance, and which component of the loss benefits most?}
\end{quote}

Clarifying this mechanism is essential for principled tokenizer design and for understanding the true drivers of scaling laws in language modelling. 

\vspace{-0.2\baselineskip}
\section{Experiments}
\vspace{-0.2\baselineskip}
\paragraph{Guiding Questions.}
Before diving into settings and metrics, we spell out the concrete questions that steer our empirical study:

\begin{enumerate}
    \item \textbf{Tokenized Text Complexity}\\
          Does enlarging the vocabulary reduce the Kolmogorov complexity of tokenized text (§\ref{main:complexity})?
    \item \textbf{Skew vs.\ Segmentation}\\
          Does enlarging the vocabulary \emph{mainly} sharpen the relative token-frequency distribution,
          or does it still increase single-token coverage of frequent words (§\ref{main:segment efficiency})?
    \item \textbf{Loss Decomposition}\\
          When the complexity decreases and frequency skew increases, how is cross-entropy re-allocated between frequent and rare tokens, and which drives the global loss (§\ref{main:token freq})?
    \item \textbf{Corpus Robustness}\\
          Are the above effects stable across different data quality—i.e.\ do high-curation (FineWeb-Edu) and
          noisier (OpenWebText) corpora exhibit the same trends (§\ref{main:dataset quality})?
\end{enumerate}

The remainder of this section answers these questions in turn.

\vspace{-0.2\baselineskip}
\subsection{Experimental Settings} \label{main:setting}
\vspace{-0.2\baselineskip}
In these experiments, we train a byte-pair encoding (BPE) tokenizer \cite{BPE} and estimate token frequencies using a sample of 10 billion GPT-2 tokens from FineWeb-Edu \cite{penedo2024finewebdatasetsdecantingweb} and the entire OpenWebText \cite{Gokaslan2019OpenWeb}. For model pre-training, we use approximately 40 billion characters, about 7.5 billion tokens for FineWeb-Edu and 7 billion for OpenWebText with a 49K vocabulary. To compute the metrics below, we also drew an additional 5 billion characters that did not overlap with the training corpus. We report word-level average loss to ensure fair comparison across vocabulary sizes. Whenever a smaller-vocabulary tokenizer splits a word into multiple tokens (i.e., subwords), we sum their individual losses. Our model comprises 85 million non-embedding parameters with pre-layer normalization (pre-LN) \cite{xiong2020layernormalizationtransformerarchitecture}. Training uses AdamW \cite{Adamw} ($\beta_1 = 0.9$, $\beta_2 = 0.95$, $\epsilon = 10^{-8}$) with a learning rate of \(6\times10^{-4}\) that follows a cosine-decay schedule after a 350 million-token warmup, weight decay of $0.1$, and gradient clipping at $1.0$. Every experiment was repeated with five seeds (See Appendix \ref{apdx:setting}).

\vspace{-0.2\baselineskip}
\subsection{Metrics} 
\vspace{-0.2\baselineskip}
\paragraph {Kolmogorov Complexity of Tokenized Text} Kolmogorov complexity $K(X)$ measures the minimal description length of a bitstring $X$ \citep{kolmogorov,goldblum2023}, thus offering an algorithmic information–theoretic way to quantify the number of statistical patterns in the data. This is different from Shannon entropy which is the lower bound of an expected code length for a random variable under its probability distribution \cite{shannon1948,englishshannonentropy}. Two concepts are highly correlated each other: shannon entropy approximates an average per-token kolmogorov complexity \cite{grunwald2004,Morris2025}. We adopt Kolmogorov complexity as a general metric for measuring complexity of tokenized text since different pre-tokenization rules can change token counts dramatically under the same tokenizer settings \cite{superbpe,schmidt2025boundlessbytepairencoding,Reddy2025}.

Exact Kolmogorov complexity is uncomputable as it reduces to the halting problem \citep{Kolmogorov1968,Morris2025,goldblum2023,kolmogorov_test}. Instead, we calculate a computable upper bound on Kolmogorov complexity, which serves as a practical metric for measuring the complexity of compressed data \citep{goldblum2023}. As BPE tokenization \cite{BPE} is derived from a BPE compression algorithm \citep{Gage1994}, tokenized text can be viewed as compressed data. For a BPE-tokenized bitstring $X^{N}$, an asymptotic upper bound on $K(X^{N})$ is given by

\vspace{-0.42\baselineskip}
\begin{equation*}
K(X^{N}) \;\le\; N\,H(p) \;+\; V \log_{2} N \;+\; O(\log N),
\end{equation*}

where \(N\) is the total token count of tokenized text, \(H(p)\) is the unigram Shannon entropy of the token distribution, and the \(V\log_{2}N\) term accounts for the prefix-free encoding of the token (e.g., token frequency table). Since each token count is an integer $\leq N$ requiring $\left\lceil \log_{2}\!\left(N+1\right) \right\rceil$ bits, the BPE model is $\leq V \log_{2} N$ bits. The first term upper-bounds the tokenized data size, the second bounds the frequency table size, and the third captures minor logarithmic overhead. Since modern language models are trained on a massive corpus with billions of tokens, the \(N\,H(p)\) component dominates. We use $K(X^{N}) \approx N\,H(p)$, focusing on the complexity of the tokenized sequence itself. Accordingly, we take $H(p)$ as the unigram Shannon entropy, despite it being a loose upper bound, as it is more stable for small datasets and alleviates data sparsity issues.


\paragraph {Loss Decomposition Metrics} In these experiments, we calculate three metrics to assess the language model’s performance both for individual vocabulary types and overall: (1) Total Loss, (2) Average Per-Word Loss, and (3) Global Cross-Entropy Loss. 

For each vocabulary \(v\), we accumulate its \emph{Total Loss}:

\begin{equation*}
\mathrm{Total\ Loss}(v)
\;=\;
\sum_{t\in N}\sum_{i=1}^{|t|}
\mathbbm{1}(v = t_i)\,\bigl[-\ln p(t_i \mid t_{<i})\bigr],
\end{equation*}

where \(N\) is the set of evaluation documents, \(\mathbbm{1}(v = t_i)\) is the indicator that the \(i\)th token equals vocabulary \(v\), and \(-\ln p(t_i\mid t_{<i})\) is the negative log-likelihood of that token. Total loss measures the sum of negative log-likelihoods for each vocabulary, reflecting its impact on the model's loss \cite{magnusson2024palomabenchmarkevaluatinglanguage}.

\emph{Average Per-Word Loss} for vocabulary \(v\) is defined as
\begin{equation*}
\mu(\ell_v)
\;=\;
\frac{\mathrm{Total\ Loss}(v)}{T_v(N)},
\end{equation*}

where  \(T_v(N)\) is the total count of occurrences of \(v\) in the training data. This represents the mean negative log-likelihood across all occurrences of each vocabulary \(v\) \cite{magnusson2024palomabenchmarkevaluatinglanguage}.

The \emph{Global Cross-Entropy Loss} is the weighted average of these per-word losses:
\begin{equation*}
\mathrm{Global\ Cross\ Entropy\ Loss}
=\sum_{v}\frac{T_v(N)}{T(N)}\,\mu(\ell_v),
\end{equation*}

where \(T(N)=\sum_v T_v(N)\) is the total token count in the training data. Global Cross-Entropy Loss reflects the model's average uncertainty per prediction \cite{Jelinek1977Perplexity, magnusson2024palomabenchmarkevaluatinglanguage}. 

\begin{wraptable}[13]{r}{0.34\columnwidth}  
\vspace{-1.0\baselineskip}
  \centering
  \caption{Upper bound of Kolmogorov complexity ($K(X^{N})$) and NCR for the $45.97$ billion byte FineWeb‑Edu corpus with $K(X^{N})$ measured in bytes.}
  \label{tab:complexity}
  \resizebox{\linewidth}{!}{%
    \begin{tabular}{lcr}
      \toprule
      \textbf{Vocab size} & \textbf{$K(X^{N})$} & \textbf{NCR} \\
      \midrule[1.5pt]
      \multirow{1}{*}{\textsc{$24\mathrm{K}$}}   & $10.74B$     & 0.234 \\
      \midrule
      \multirow{1}{*}{\textsc{$49\mathrm{K}$}}  & $10.43B$     & 0.227 \\
      \midrule
      \multirow{1}{*}{\textsc{$98\mathrm{K}$}}    & $10.23B$     & 0.223 \\
      \midrule
      \multirow{1}{*}{\textsc{$196\mathrm{K}$}}    & $10.16B$     & 0.221 \\
      \bottomrule
    \end{tabular}%
  }
\end{wraptable}

\subsection{Increasing vocabulary size reduces complexity of tokenized text} \label{main:complexity}

Text complexity can be viewed as the amount of statistical patterns in the data, where such patterns refer to regularities that appear repeatedly in text, such as word frequency, contextual co-occurrence, and sentence structure. To quantify how larger vocabulary size reduces tokenized text complexity, we measure the upper bound of Kolmogorov complexity and the normalized compression ratio (NCR). NCR is a practical compressibility metric $\mathrm{NCR}(x; C) \;=\; \frac{\lvert C(x)\rvert}{\lvert x\rvert}$ where \(\lvert x\rvert\) is the byte length of data \(x\) and \(\lvert C(x)\rvert\) is its size after lossless compression by compressor \(C\); in our formulation, $K(X^{N}) \approx \lvert C(x)\rvert = N\,H(p)$ where $N$ is token count and $H(p)$ is unigram Shannon entropy.  


Table \ref{tab:complexity} reports the upper bound on Kolmogorov complexity and NCR for the $45.97$ billion byte FineWeb‑Edu corpus ($10\mathrm{B}$ GPT-2 tokens) \cite{penedo2024finewebdatasetsdecantingweb}. The results show that BPE tokenizer with larger vocabularies yields lower complexity and NCR by segmenting frequent non-i.i.d. character sequences in text (e.g., words) as a single token. This can be interpreted as larger vocabularies reduce the number of statistical patterns that models need to learn by offloading the low-level pattern learning, thereby simplifying language modeling.

\begin{figure}[t]
\vskip -0.1in
  \centering
  \begin{subfigure}[b]{0.32\textwidth}
    \centering
    \includegraphics[width=\textwidth]{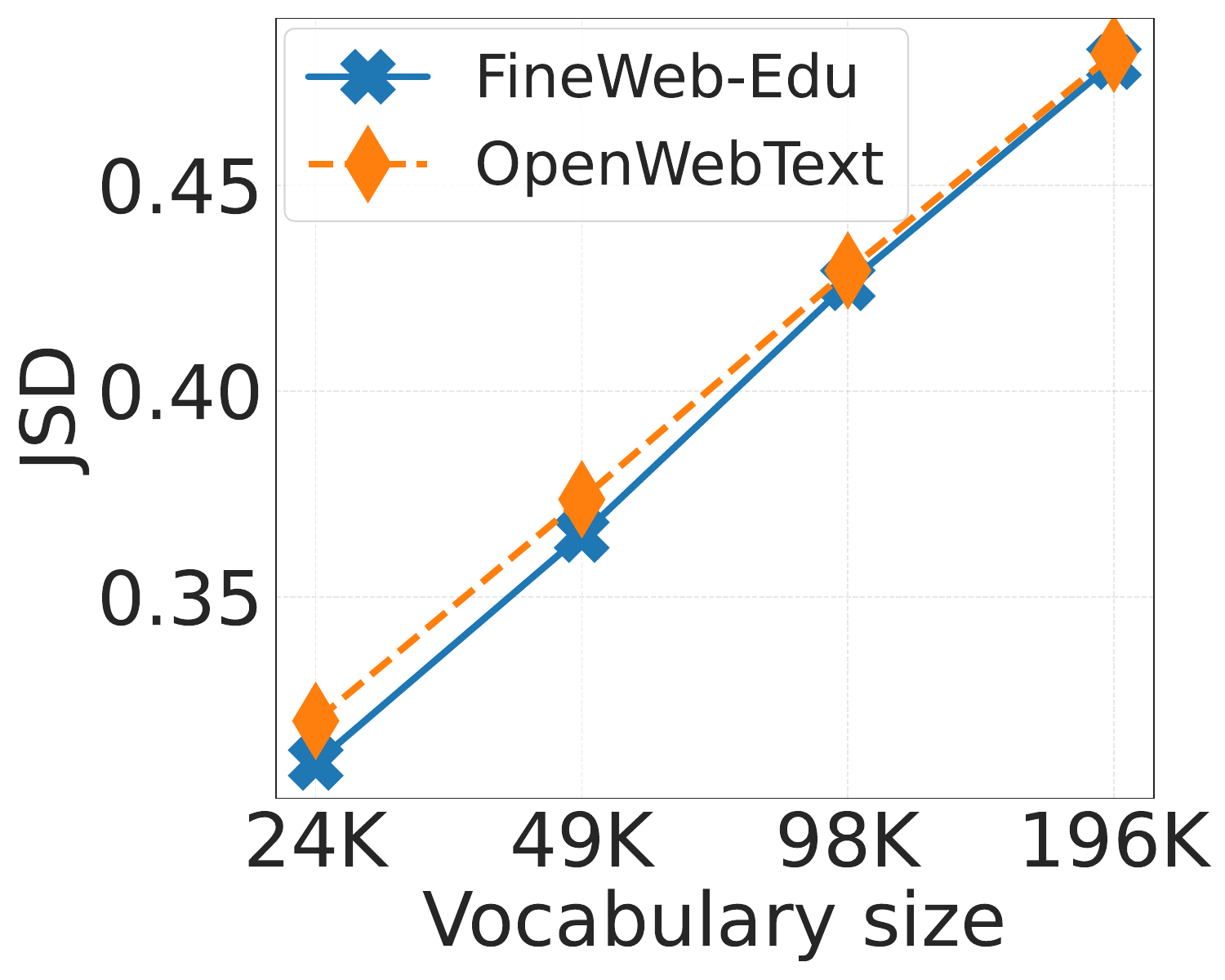}
    \caption{Token-frequency imbalance}
    \label{fig:2a}
  \end{subfigure}%
  \hfill
  \begin{subfigure}[b]{0.32\textwidth}
    \centering
    \includegraphics[width=\textwidth]{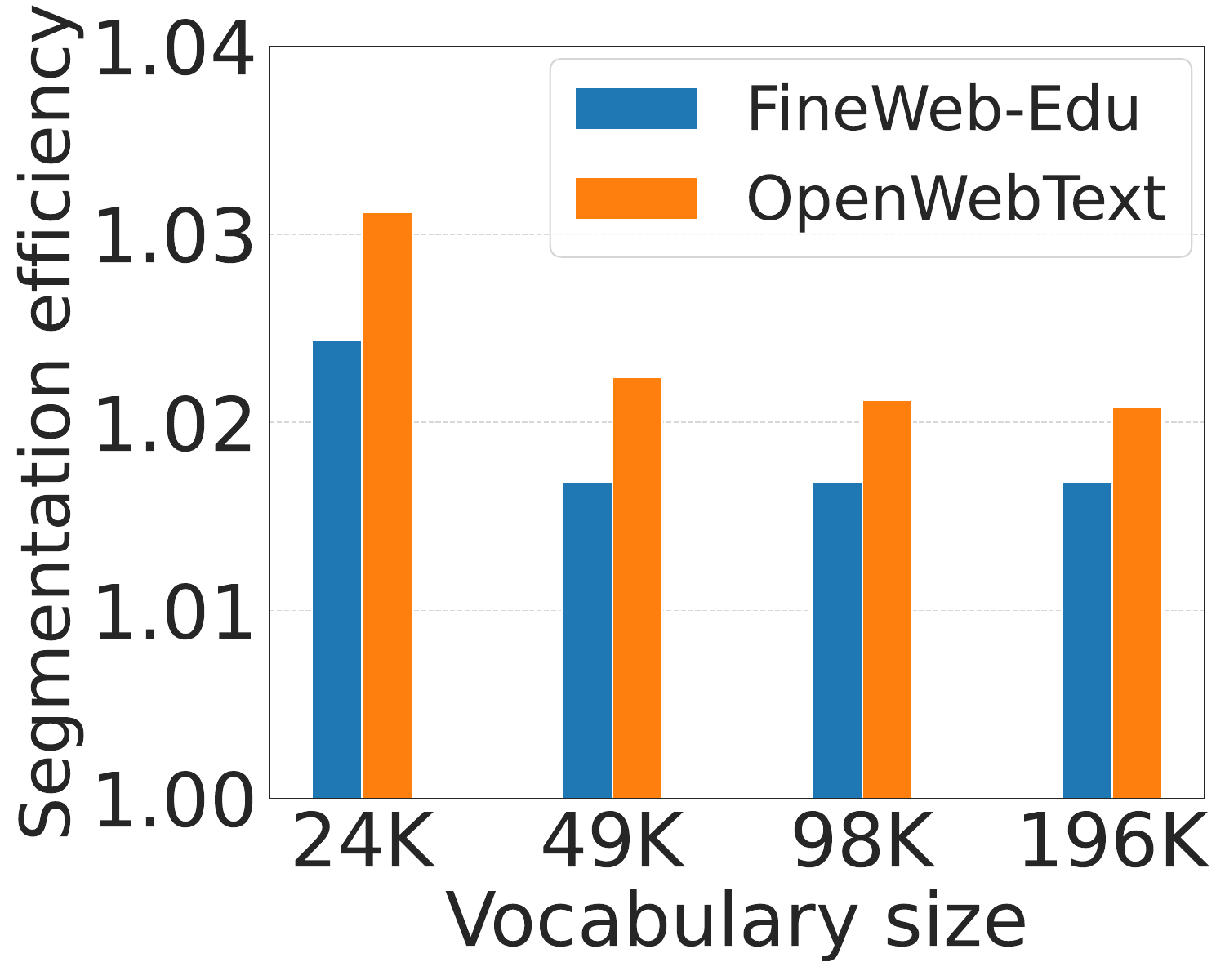}
    \caption{Number of tokens per word}
    \label{fig:2b}
  \end{subfigure}%
  \hfill
  \begin{subfigure}[b]{0.32\textwidth}
    \centering
    \includegraphics[width=\textwidth]{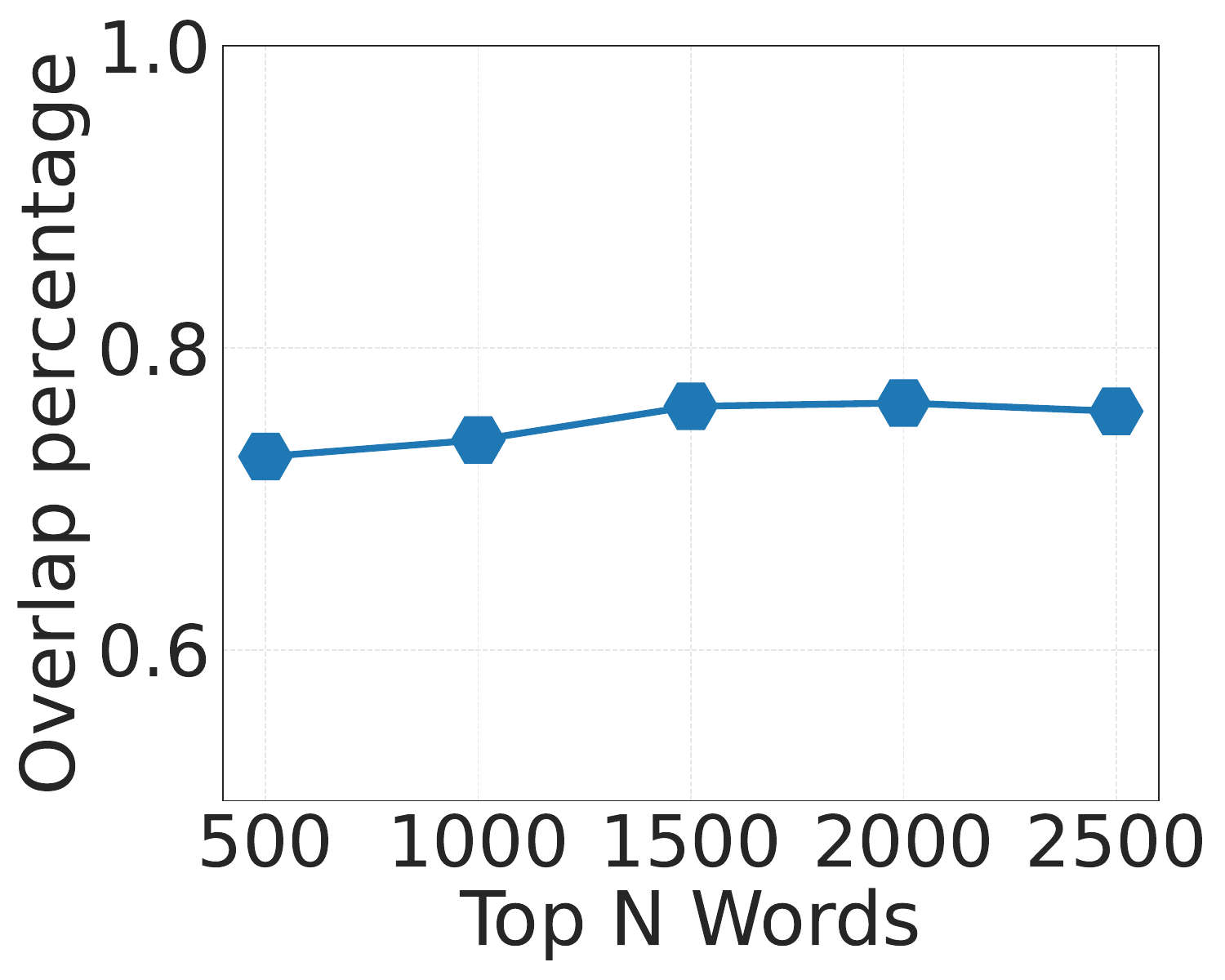}
    \caption{Overlap of frequent N words}
    \label{fig:2c}
  \end{subfigure}

  \caption{Figure \ref{fig:2a} shows that increasing vocabulary size exacerbates relative token-frequency imbalance. In other words, enlarging the vocabulary size introduces more rare tokens, causing the relative token-frequency distribution to be further from a uniform distribution. Figure \ref{fig:2b} reveals that a $24$K vocabulary size tokenizer already segments $2,500$ frequent words as a single token regardless of dataset quality. This implies that further vocabulary growth offers no added benefit for estimating the probabilities of frequent words. Figure \ref{fig:2c} shows that the most frequent $n$ words in fineWeb-Edu and OpenWebText largely overlap, highlighting the universality of frequent vocabulary across different datasets. We report the most frequent $2,500$ words in  FineWeb-Edu and OpenWebText, which account for approximately $74.4\%$ and $75.5\%$ of each dataset, respectively. } 
  \label{fig:Dataset skewness and overlap}
  \vskip -0.1in
\end{figure}

\subsection{Segmentation already saturates; vocabulary growth mainly sharpens frequency skew}\label{main:segment efficiency}
To quantify how each factor evolves with vocabulary growth, we disentangle two factors: (i) relative token-frequency imbalance (Figure~\ref{fig:2a}) and (ii) segmentation efficiency (Figure~\ref{fig:2b})—the fraction of frequent words that are represented by exactly one token. \textit{Frequent words} are operationally defined as the top $2,500$ word types, chosen because this cutoff already covers $\ge 70\%$ of corpus tokens in both FineWeb-Edu ($74.4$ \%) and OpenWebText ($75.5$ \%; full coverage curves in Appendix~\ref{apdx:coverage}).

In this experiment, relative token-frequency imbalance is quantified by the Jensen–Shannon divergence (JSD) from a uniform distribution of the same vocabulary size for a fair comparison across different vocabulary sizes \cite{zouhar_tokenization,dagan2024gettingtokenizerpretrainingdomain,schmidt2025boundlessbytepairencoding}. Segmentation efficiency gauges the average token count per word, that is, how many tokens the tokenizer needs, on average, to encode each of the $2,500$ most frequent words.
Figure~\ref{fig:2a} shows that JSD grows monotonically with vocabulary size, reflecting greater token-frequency imbalance. In contrast, Figure~\ref{fig:2b} indicates that the segmentation efficiency exceeds $95$ \% by $24\mathrm{K}$ tokens and plateaus later. In other words, beyond $24\mathrm{K}$, the vocabulary gets larger \textit{without} providing additional single-token coverage for frequent words. Within the widely used $24\mathrm{K}\!-\!196\mathrm{K}$ range \citep{schmidt2024tokenizationcompression, ali2024tokenizerchoicellmtraining, reddy2025enoughdiminishingreturnstokenization}, enlarging the vocabulary chiefly amplifies relative token-frequency imbalance rather than improving segmentation efficiency. This finding tempers the common view that "bigger vocabularies help by approximating word-level tokens" \citep{rajaraman2025theorytokenizationllms}: that mechanism appears to saturate once frequent words are already tokenized as a single token.

\subsection{Skew-driven effects persist across corpora quality levels} \label{main:dataset quality} 

To test whether the effects of increasing vocabulary size hold across datasets of varying quality, we conduct experiments on both FineWeb-Edu \cite{penedo2024finewebdatasetsdecantingweb} and OpenWebText \cite{Gokaslan2019OpenWeb}. 
Figure \ref{fig:2a} and \ref{fig:2b} demonstrate that the impact of increasing vocabulary size has a similar effect on both high-quality datasets (e.g., FineWeb-Edu \cite{penedo2024finewebdatasetsdecantingweb}) and the lower-quality ones (e.g., OpenWebText \cite{Gokaslan2019OpenWeb}). Figure \ref{fig:2c} further shows that the frequent $2,500$ words coincide with nearly $75$\% each other, highlighting the universality of frequent vocabulary across different corpora. Overall, the findings indicate that vocabulary size expansion produces the same effects irrespective of corpus quality and that high‑frequency words substantially overlap across datasets.

\subsection {Larger vocabularies reduce frequent words uncertainty and global loss} \label{main:token freq}

\begin{figure}[t]
\vskip -0.1in
  \centering
  \begin{subfigure}[b]{0.49\textwidth}
    \centering
    \includegraphics[width=\textwidth]{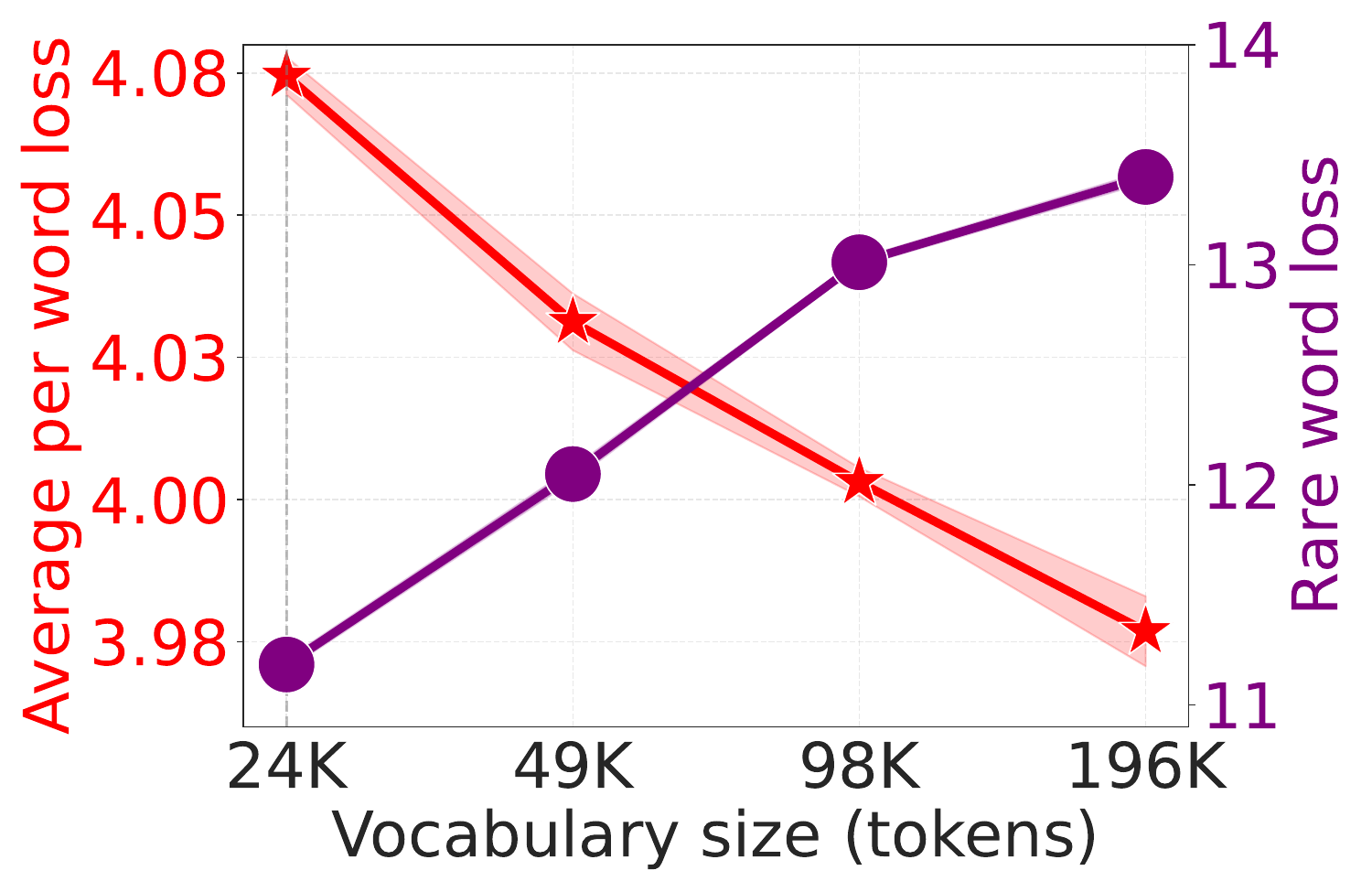}
    \caption{Frequent vs. rare word loss (85M, 10B)}
    \label{fig:3a}
  \end{subfigure}%
  \hfill
  \begin{subfigure}[b]{0.49\textwidth}
    \centering
    \includegraphics[width=\textwidth]{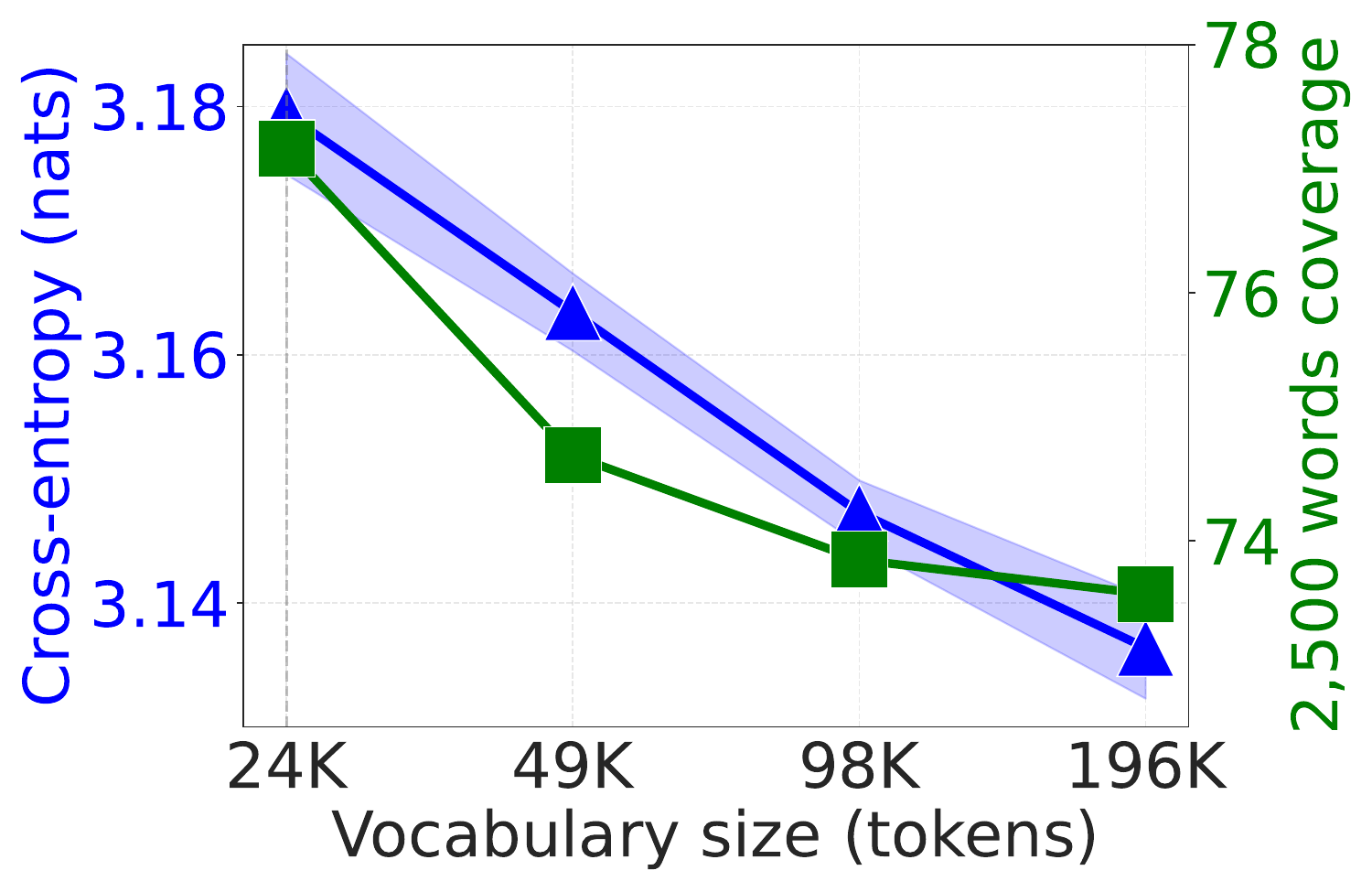}
    \caption{CE-loss \& frequent-word coverage (85M, 10B)}
    \label{fig:3b}
  \end{subfigure}%

  \caption{Figure \ref{fig:3a} illustrates that models with a larger vocabulary size reduce loss on the most frequent $2,500$ words while increase loss on the rarest $20,000$ words. Nevertheless, Figure \ref{fig:3b} shows that the global cross‐entropy loss declines as vocabulary size increases, demonstrating that the gains from lower loss on frequent words outweigh the losses from poorer infrequent word estimates. It further reveals that frequent words account for nearly $75$\% of the total loss, while loss on infrequent words grows with vocabulary size as their conditional probabilities fall due to data sparsity. Models are pre-trained on $40\mathrm{B}$ bytes and evaluated on a disjoint $5\mathrm{B}$ byte split of FineWeb-Edu.}
  \label{fig:scaling vocab}
  \vskip -0.1in
\end{figure}

The effect of tokenized text complexity and relative token-frequency imbalance on language models has not yet been explored. Each prediction in a language model depends on previous tokens, and the conditional nature of these probabilities adds significant analytical complexity. However, it is possible to hypothesize that loss of rare token prediction weighs heavily in the overall loss, as conditional probability is often orders of magnitude lower and therefore yields much higher per‑token losses \cite{pinto2024fairlanguagemodelparadox, mircea2024gradient}. To address this presumption, we empirically examine how reducing loss on frequent word prediction at the expense of rare word predictions affects overall model performance. Models are pre-trained on $40\mathrm{B}$ bytes and evaluated on a separate, non-overlapping $5\mathrm{B}$ byte split of FineWeb-Edu. Figure \ref{fig:3a} shows that enlarging the vocabulary steadily lowers the average per‑word loss for the frequent words, where the gap between $24\mathrm{K}$ and $196\mathrm{K}$ is roughly $0.1$ nats for the frequent $2,500$ words. In contrast, the average per-word loss for the rarest $20,000$ words rises with vocabulary size, increasing from roughly $11.183$ to $13.399$ nats. As shown in Figure \ref{fig:3b}, the global cross-entropy declines from about $3.179$ to $3.136$ nats, implying that gains on frequent-word prediction outweigh the degradation on infrequent words. Frequent tokens dominate the objective in all settings: the top $2{,}500$ words account for nearly $75\%$ of total loss. Nevertheless, as the vocabulary expands, losses on infrequent tokens grow, reflecting lower conditional probabilities and reduced predictability for those items. These observations highlight a key takeaway: once loss on frequent words declines, the overall objective is governed by their loss contribution, so further skewness in the token frequency distribution generally improves performance even though it harms rare-token estimates. This finding is consistent with the existing study \cite{rajaraman2025theorytokenizationllms, magnusson2024palomabenchmarkevaluatinglanguage}, explaining the benefits of increasing vocabulary size. We observe the same qualitative pattern in the OpenWebText (Appendix \ref{apdx:openwebtext}).

\subsection {Gains from vocabulary expansion hold for larger datasets and models} \label{main:30B}

\begin{figure}[t]
\vskip -0.1in
  \centering
  \begin{subfigure}[b]{0.49\textwidth}
    \centering
    \includegraphics[width=\textwidth]{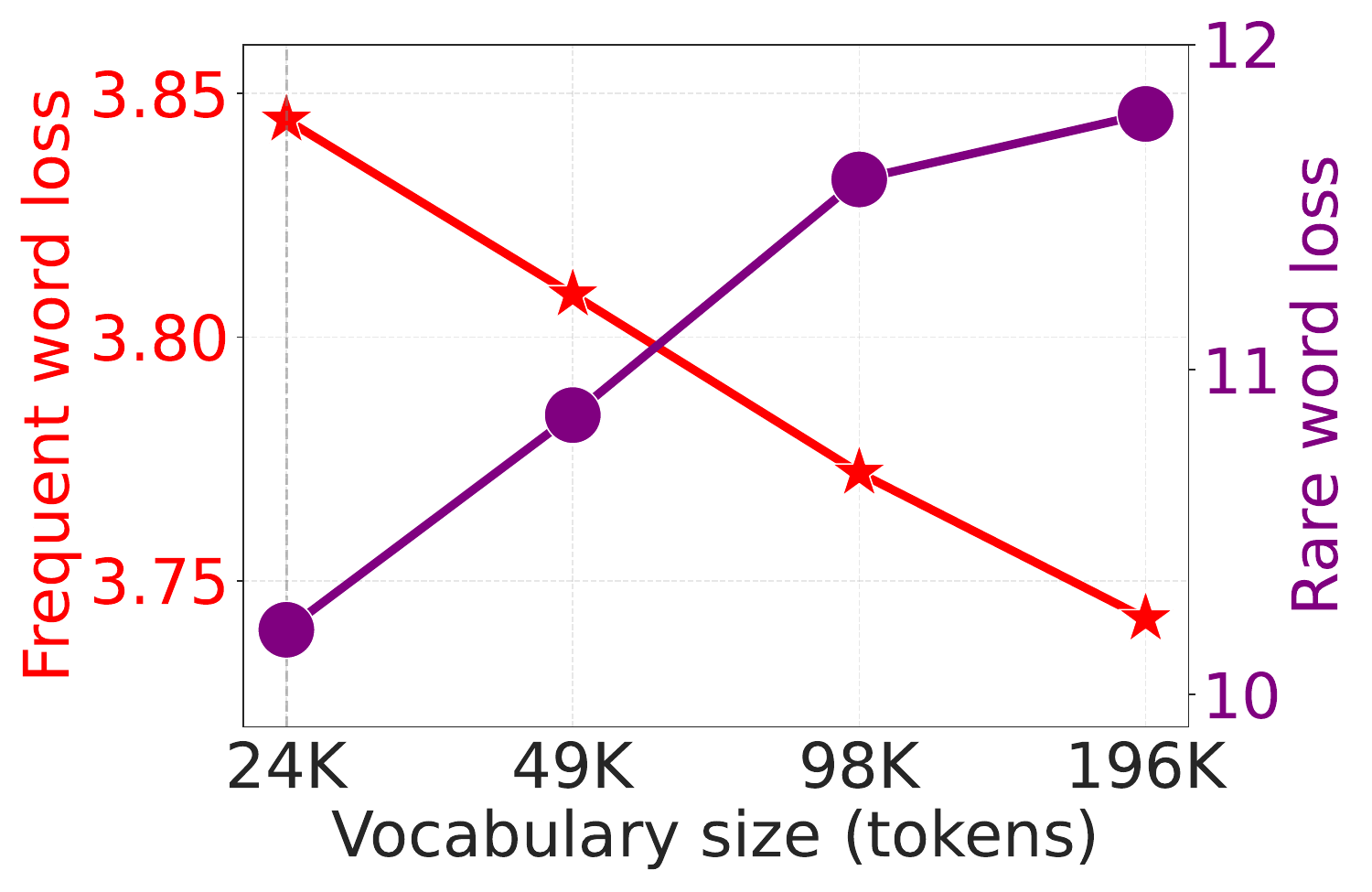}
    \caption{Frequent vs. rare word loss (85M, 30B)}
    \label{fig:30a}
  \end{subfigure}%
  \hfill
  \begin{subfigure}[b]{0.49\textwidth}
    \centering
    \includegraphics[width=\textwidth]{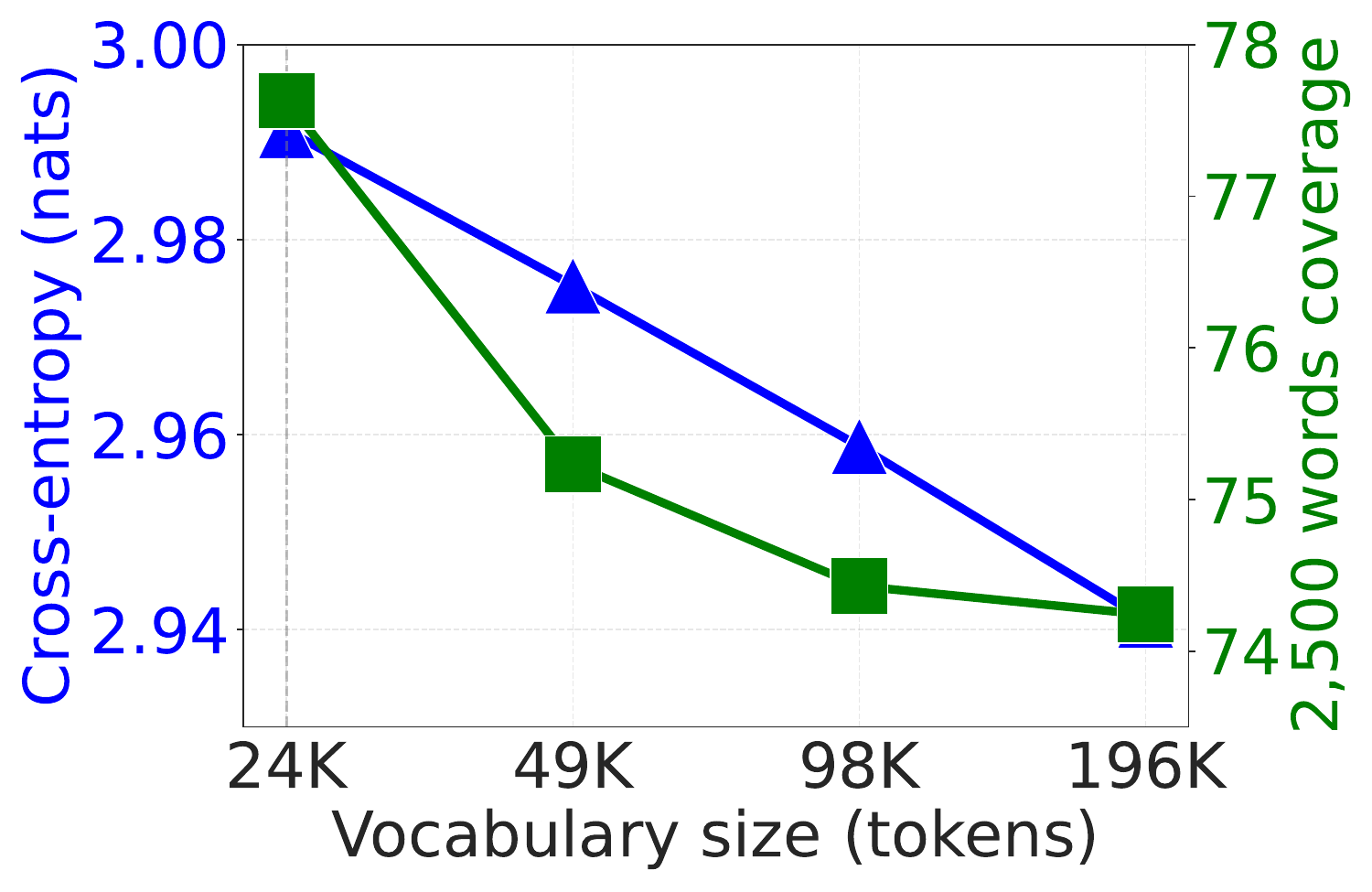}
    \caption{CE-loss \& frequent-word coverage (85M, 30B)}
    \label{fig:30b}
  \end{subfigure}%
  \hfill
  \begin{subfigure}[b]{0.49\textwidth}
    \centering
    \includegraphics[width=\textwidth]{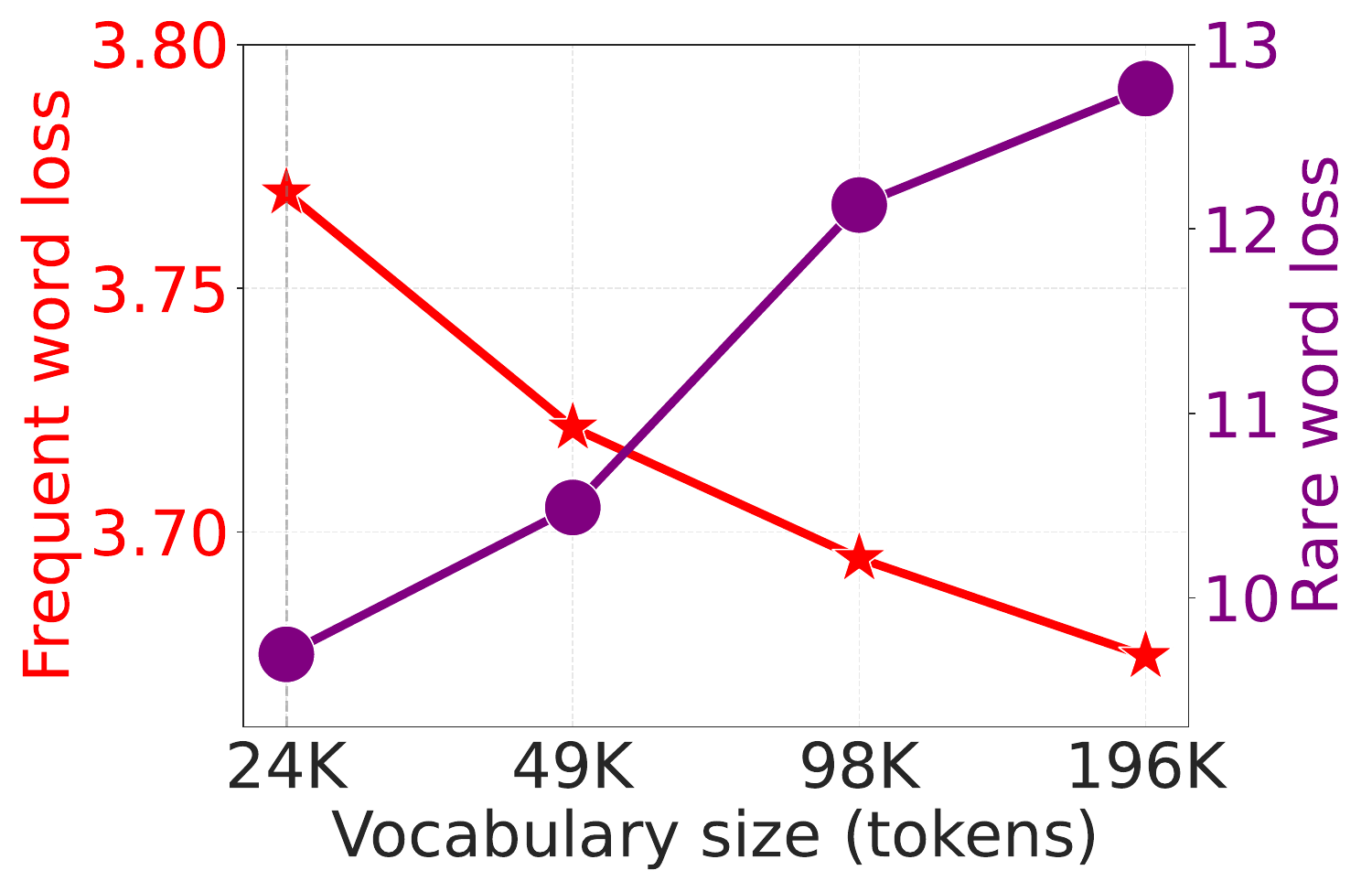}
    \caption{Frequent vs. rare word loss (450M, 10B)}
    \label{fig:30c}
  \end{subfigure}%
  \hfill
  \begin{subfigure}[b]{0.49\textwidth}
    \centering
    \includegraphics[width=\textwidth]{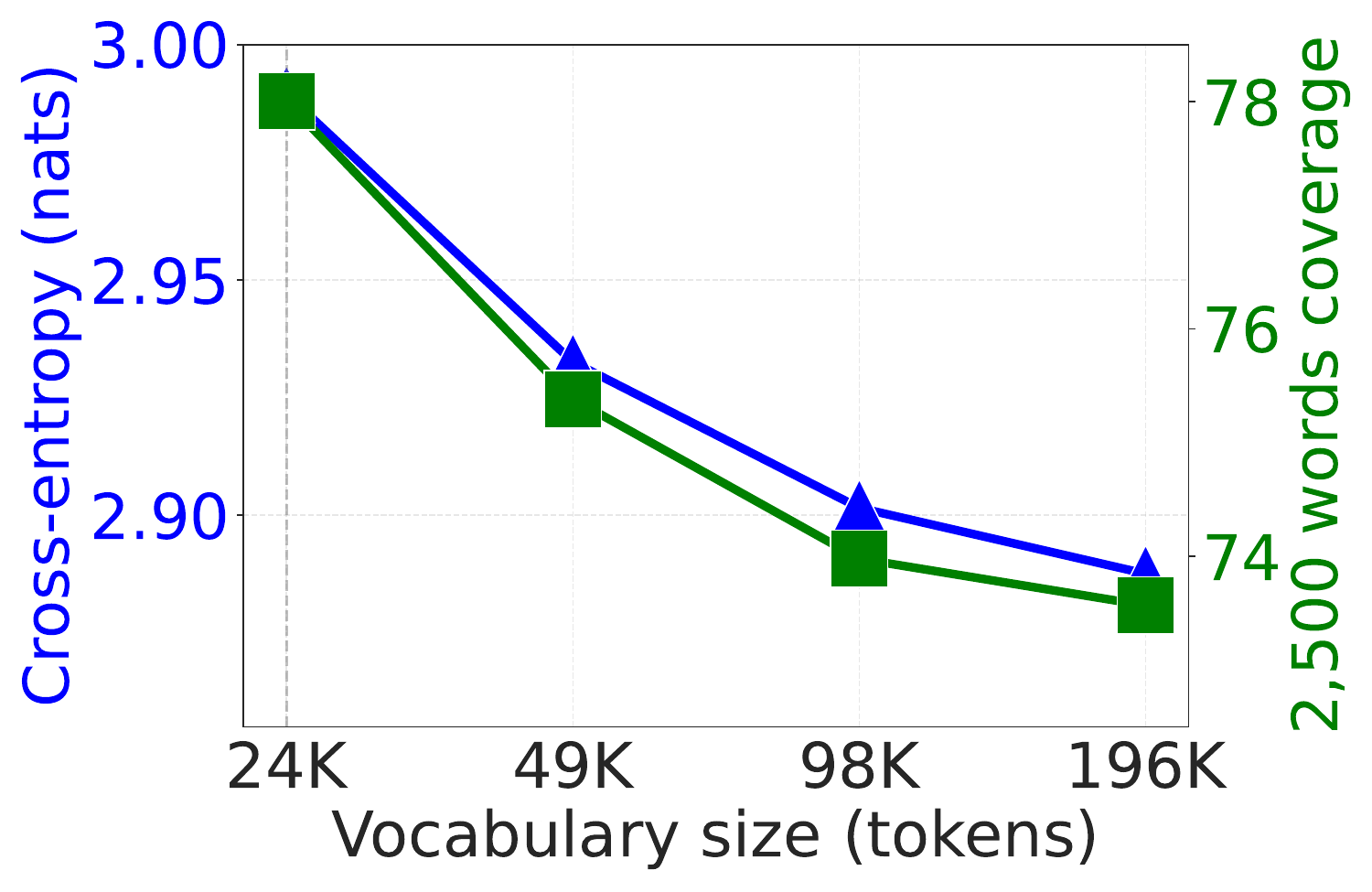}
    \caption{CE-loss \& frequent-word coverage (450M, 10B)}
    \label{fig:30d}  
  \end{subfigure}%

  \caption{For an $85\mathrm{M}$ model trained on $30\mathrm{B}$ tokens, larger vocabularies reduce the most frequent $2,500$ word loss while increase the rarest $20,000$ word loss; since frequent words dominate, global cross-entropy drops (figure \ref{fig:30a} and \ref{fig:30b}). $450\mathrm{M}$ model trained on 10B tokens mirrors the pattern (figure \ref{fig:30c} and \ref{fig:30d}), indicating that these vocabulary-size effects persist across larger datasets and models.}
  \label{fig:scaling 30B}
  \vskip -0.1in
\end{figure}

To assess whether vocabulary size effects persist across data and model scales, we train an $85\mathrm{M}$ (non-embedding) model on a $30\mathrm{B}$ GPT-2 token subset ($184.6\mathrm{B}$ bytes) of FineWeb-edu and a $450\mathrm{M}$ (non-embedding) model on a $10\mathrm{B}$ token subset. On the $30\mathrm{B}$ token split, figure \ref{fig:30a} shows frequent-word loss drops from $3.845$ to $3.742$ nats, while rare-word loss rises from $10.199$ to $11.787$ nats; figure \ref{fig:30b} shows global cross-entropy falls from $2.991$ to $2.941$ nats and frequent-word loss coverage declines from $77.6\%$ to $74.2\%$, exhibit the same trend reported in Section \ref{main:token freq}. Figure \ref{fig:30c} and \ref{fig:30d} confirm the same pattern for the 450M models: frequent-word loss decreases from $3.770$ to $3.675$ nats, rare-word loss increases from $9.694$ to $12.762$ nats, cross-entropy reduces from $2.989$ to $2.888$ nats, and frequent word loss coverage from $78.0\%$ to $73.6\%$. Overall, these findings indicate that vocabulary expansion produces the same effects with larger dataset and model size.

\section{Analysis}
Although our results show that minimizing loss on high frequency tokens is crucial for reducing global cross-entropy, several questions remain unresolved. To pinpoint the causal chain from tokenizer design to model behavior, we now organize our analysis around two additional guiding questions. 

\paragraph{Guiding Questions.}
\vskip -0.1in
\begin{enumerate}
  \item \textbf{Transfer Mechanism}\\
        How does the heavy overlap of the most frequent $2,500$ words connect pre-training loss drops to downstream accuracy (§\ref{main:generalization})?
  \item \textbf{Parameter Count vs.\ Vocabulary Scaling}\\
        Can enlarging model size (with a fixed tokenizer) reproduce the same advantage delivered by a larger vocabulary (§\ref{main:model size})?
\end{enumerate}

The next two subsections answer newly added Q$1$–Q$2$ in turn.


\subsection{Frequent-word overlap explains downstream performance transfer} \label{main:generalization}

\newlength\commonfigheight
\setlength\commonfigheight{0.25\textheight}

\begin{figure}[t]
  \centering
  \begin{subfigure}[b]{0.305\textwidth}
    \centering
    \includegraphics[width=\textwidth, height=\commonfigheight,keepaspectratio]{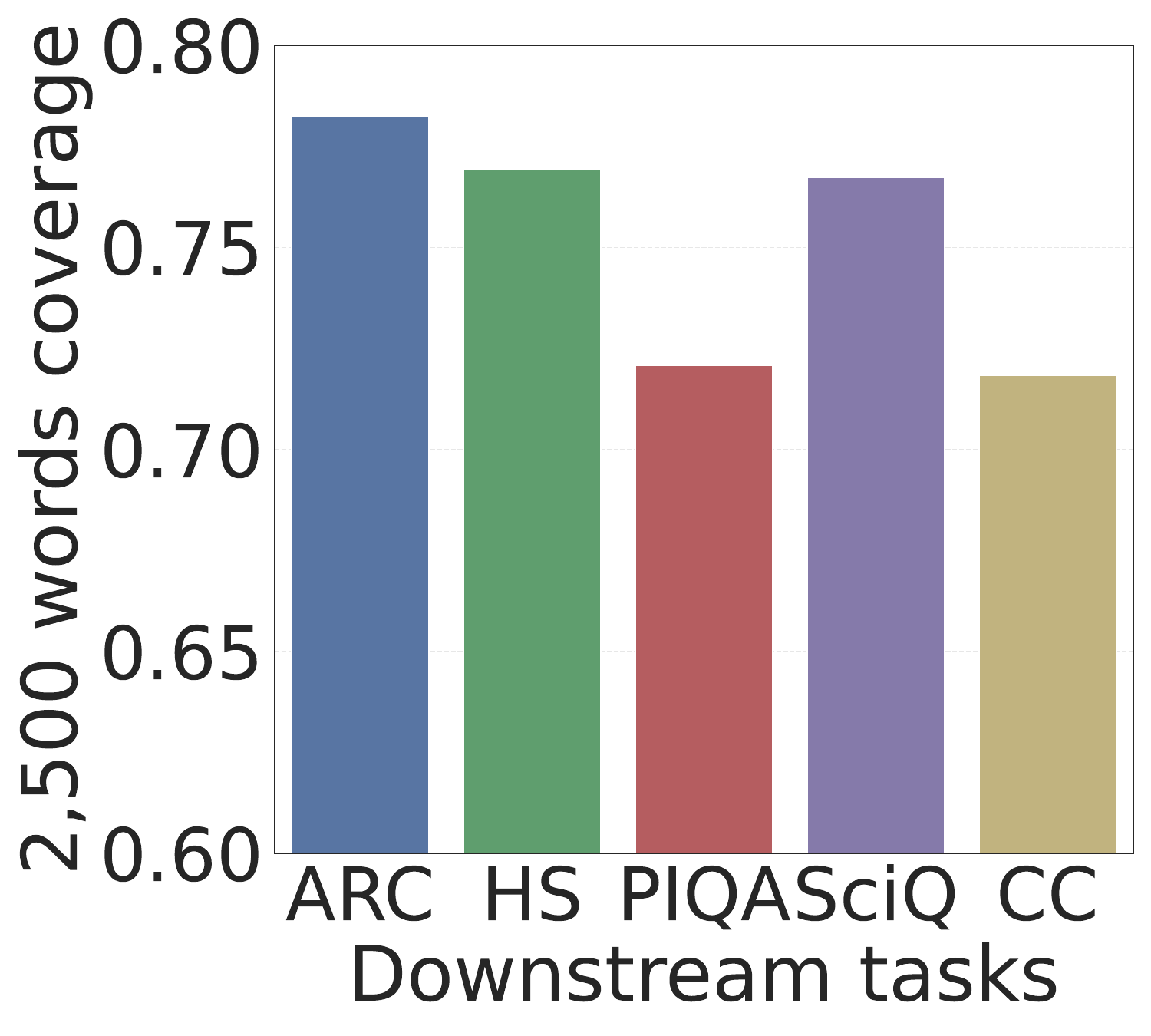}
    \caption{Frequent word coverage}
    \label{fig:4a}
  \end{subfigure}%
  \hfill
  \begin{subfigure}[b]{0.362\textwidth}
    \centering
    \includegraphics[width=\textwidth, height=\commonfigheight,keepaspectratio]{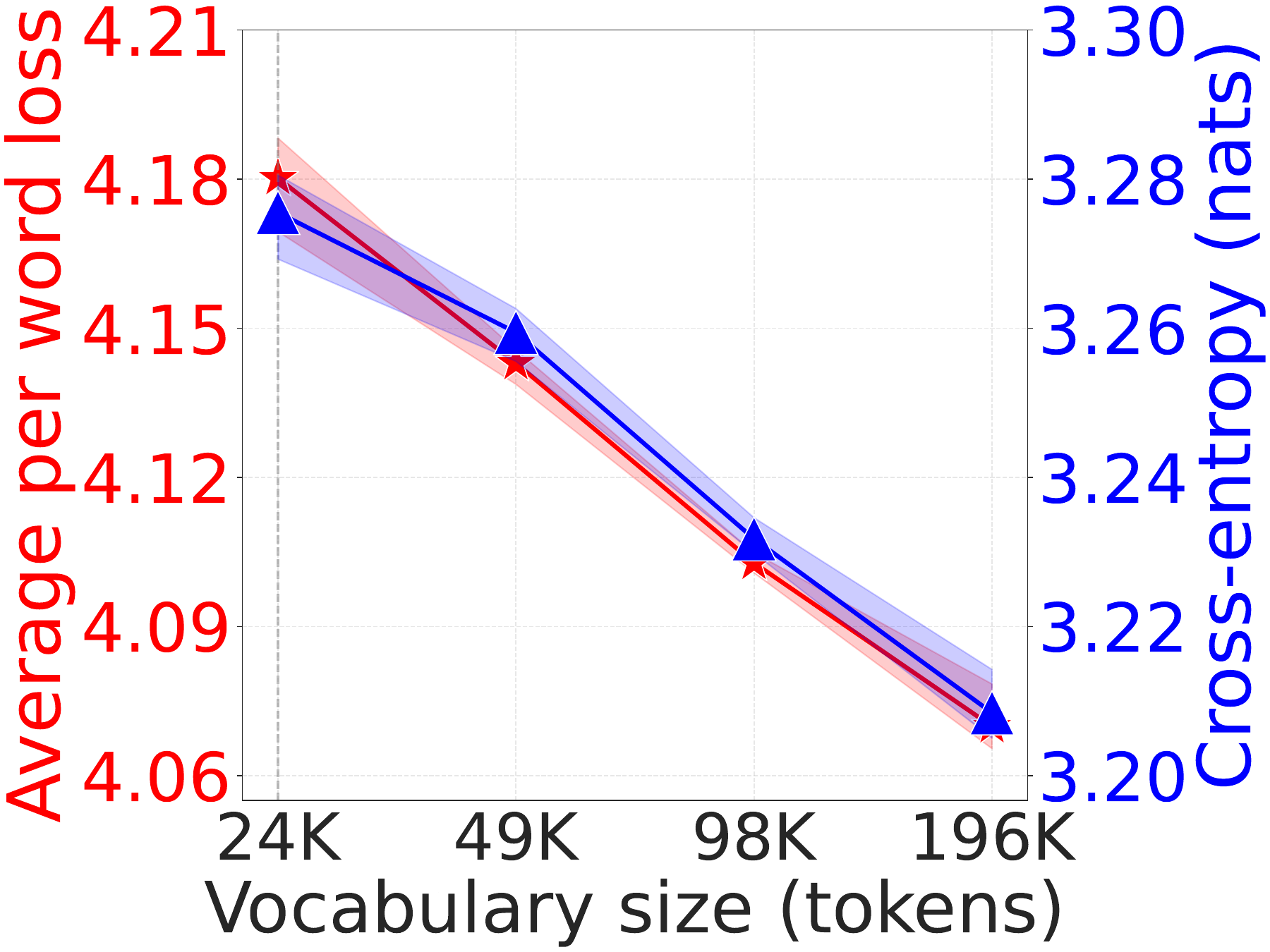}
    \caption{Frequent word loss reduction}
    \label{fig:4b}
  \end{subfigure}%
  \hfill
  \begin{subfigure}[b]{0.295\textwidth}
    \centering
    \includegraphics[width=\textwidth, height=\commonfigheight,keepaspectratio]{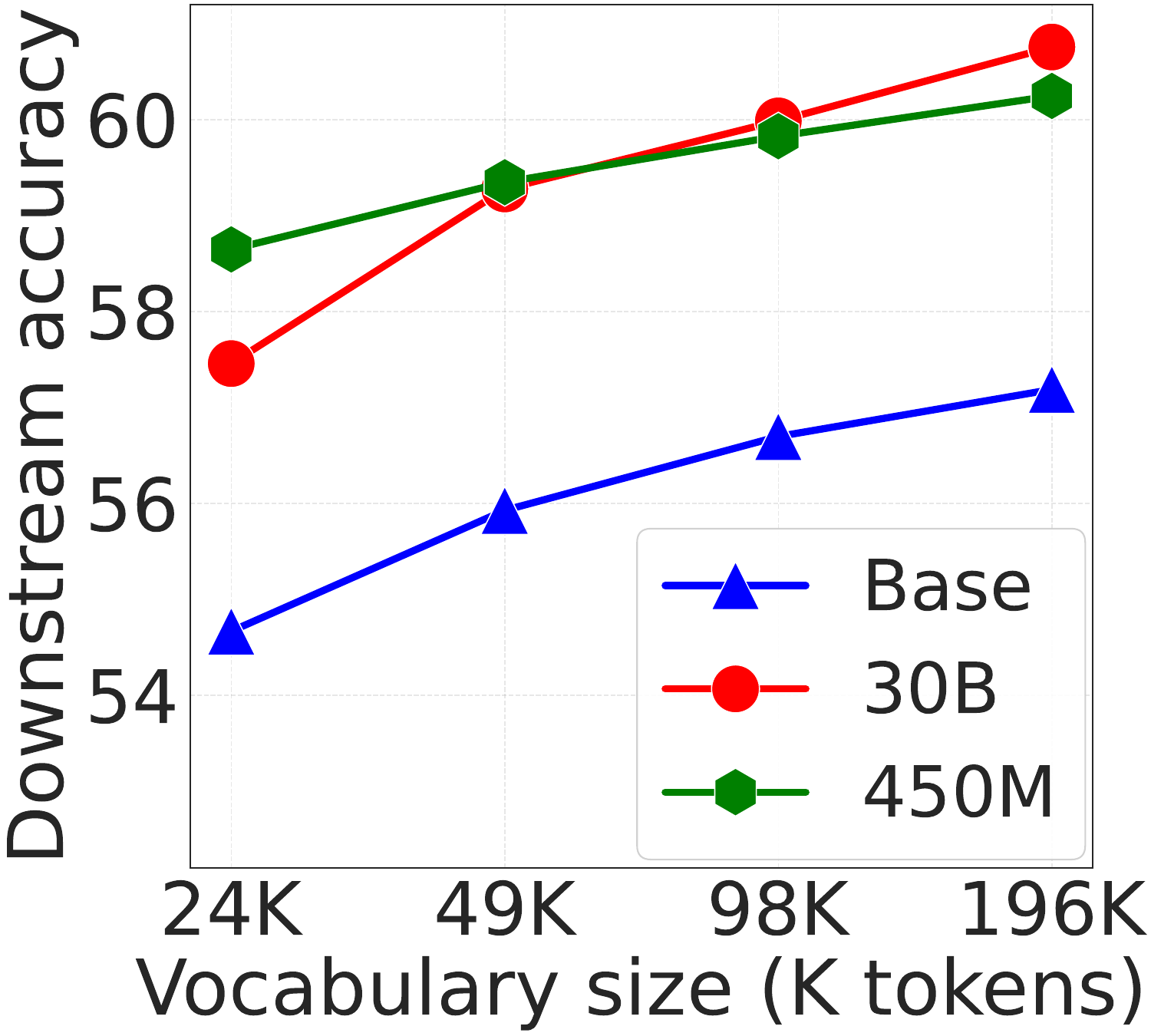}
    \caption{Downstream performance}
    \label{fig:4c}
  \end{subfigure}%
  \caption{Figure \ref{fig:4a} demonstrates that the most frequent $2,500$ words in the FineWeb-Edu comprise nearly $72-78$\% of the tokens in other downstream benchmark datasets as well as the CC-Main-$2023-40$ \cite{huang2024compressionrepresentsintelligencelinearly}. ARC refers to ARC-Easy, and HS refers to HellaSwag. Figure \ref{fig:4b} illustrates that a larger vocabulary reduces average per-word loss on frequent FineWeb-Edu words within the CC dataset, and demonstrates how this translates into lower global cross-entropy loss on CC dataset. Figure \ref{fig:4c} confirms that scaling the vocabulary size boosts downstream task performance.}
  \label{fig:generalization}
\end{figure}

Reducing loss for frequent words is key to achieving a lower global cross-entropy loss. But does this effect carry over to downstream task accuracy? Although a strong link between lower loss and better downstream performance has been studied \cite{huang2025overtokenizedtransformervocabularygenerally, kim2025perilnrevisitinglayernormalization}, its rationale has not been investigated. To shed light on this phenomenon, we analyze the overlap of frequent words between the pre-training data and evaluation benchmarks and demonstrate that scaling the vocabulary size reduces cross-entropy loss during pre-training and on downstream tasks. Figure \ref{fig:4a} shows that the $2,500$ most common words in FineWeb-Edu \cite{penedo2024finewebdatasetsdecantingweb} account for roughly $76$-$78$\% of all tokens in ARC \cite{clark2018thinksolvedquestionanswering}, HellaSwag \cite{zellers2019hellaswagmachinereallyfinish}, and SciQ \cite{sciq} and about $72$\% in the PIQA \cite{bisk2019piqareasoningphysicalcommonsense} and CC‑Main‑$2023‑40$ dump \cite{huang2024compressionrepresentsintelligencelinearly}. Because lower cross‑entropy loss on this CC subset correlates closely with stronger reasoning benchmark scores \cite{huang2024compressionrepresentsintelligencelinearly}, reducing the frequent words loss and driving down global cross-entropy will improve accuracy across downstream tasks. Consistent with that expectation, figure \ref{fig:4b} shows that increasing the vocabulary size from $24\mathrm{K}$ to $196\mathrm{K}$ lowers the average per-word loss on the frequent $2,500$ FineWeb-Edu words by about $0.11$ nats, translating into a roughly $0.07$ nats drop in global cross-entropy loss on the CC dataset. Figure \ref{fig:4c} further confirms that a larger vocabulary also improves average downstream task accuracy of the language model across datasets and model scales. The key insight from this observation is that frequent words highly overlap between the typical training dataset and the downstream benchmark, so the cross‑entropy reduction achieved by enlarging the vocabulary size during pre‑training naturally translates into better downstream performance.

\begin{figure}[t]
  \centering
  \begin{subfigure}[b]{0.48\textwidth}
    \centering
    \includegraphics[width=\textwidth]{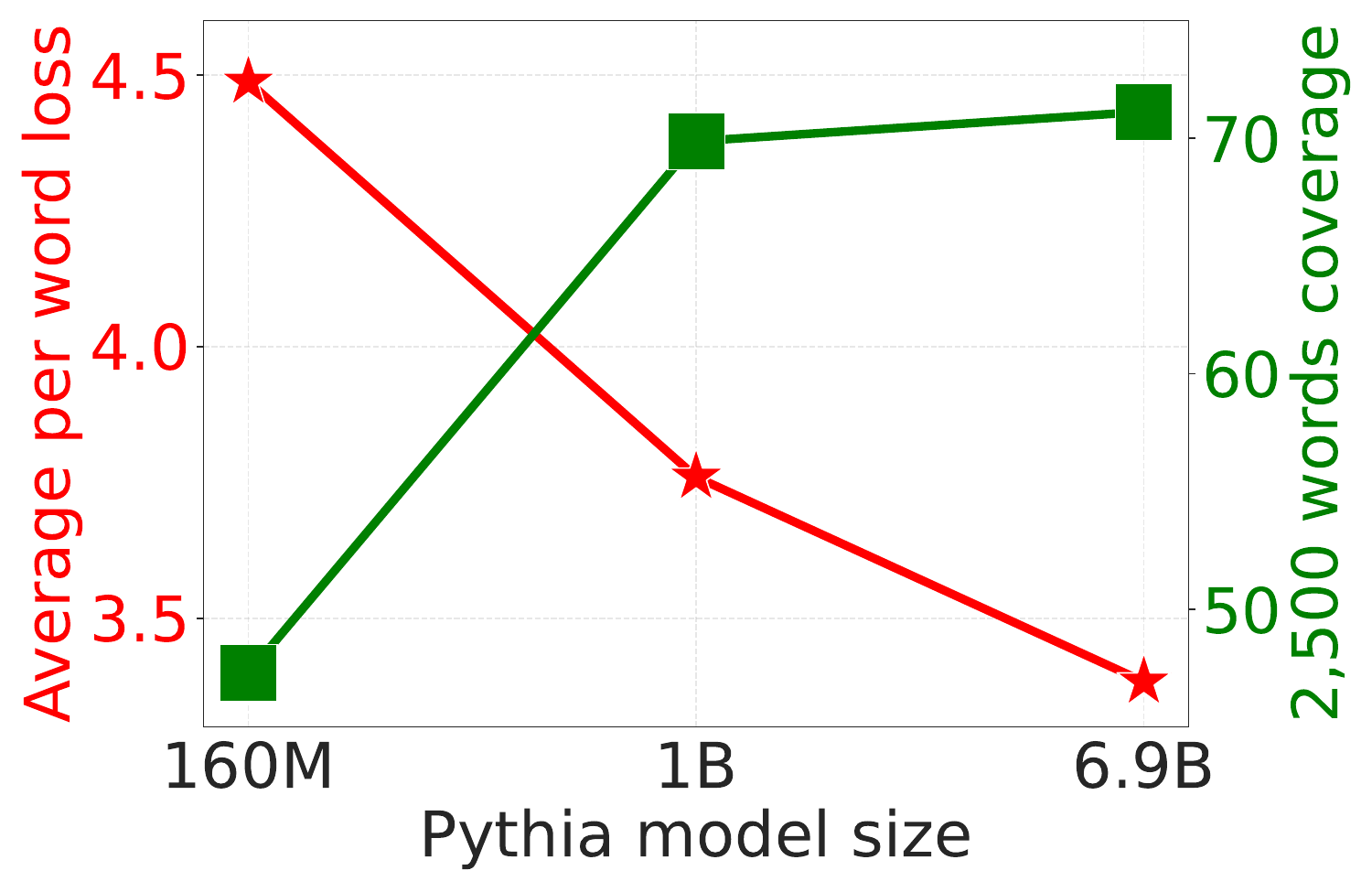}
    \caption{Frequent word loss in pythia models}
    \label{fig:5a}
  \end{subfigure}
  \hfill
  \begin{subfigure}[b]{0.48\textwidth}
    \centering
    \includegraphics[width=\textwidth]{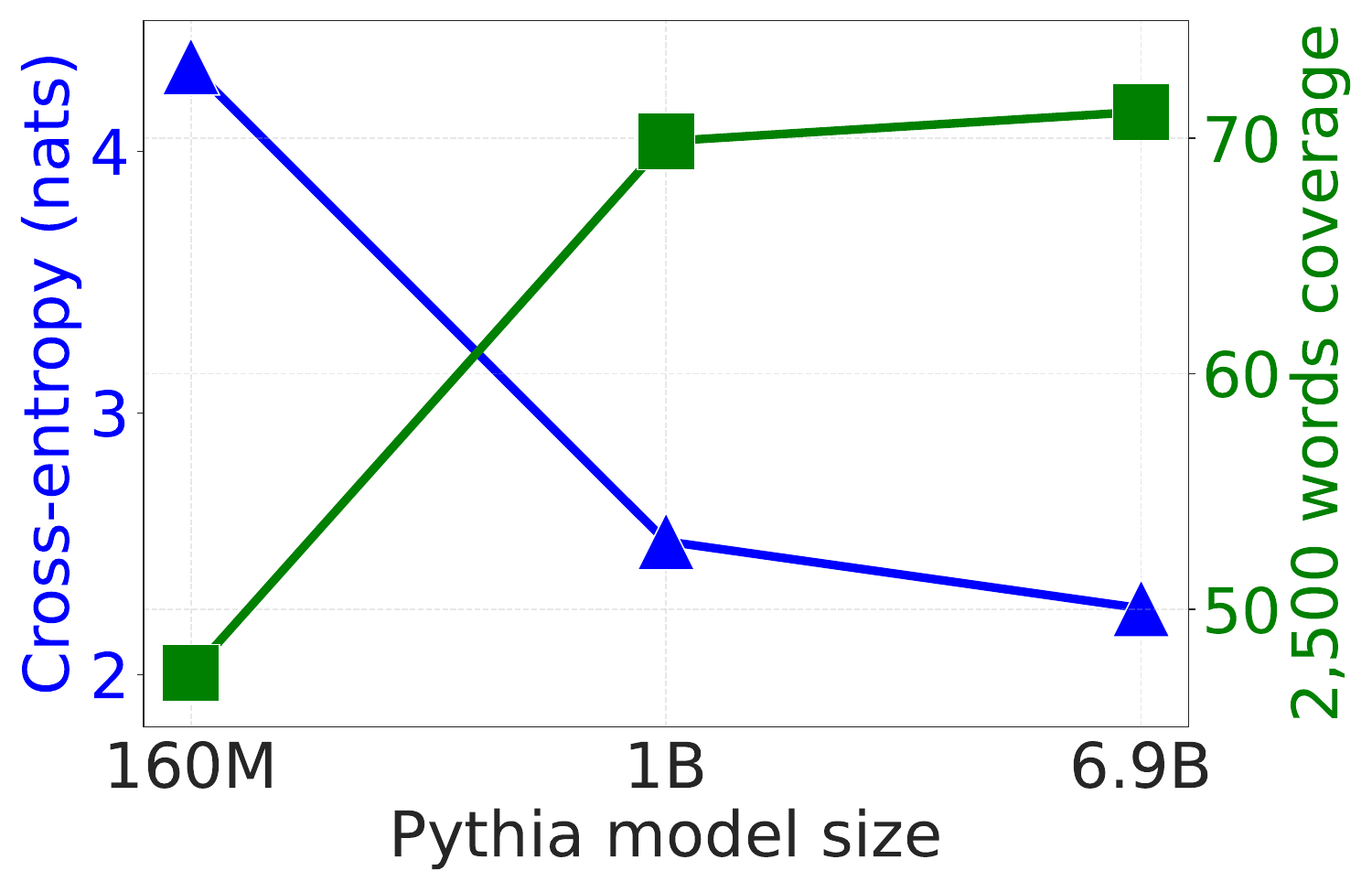}
    \caption{Global cross-entropy in pythia models}
    \label{fig:5b}
  \end{subfigure}
  \caption{Figure \ref{fig:5a} illustrates that increasing model size reduces loss on high frequency words, and the global cross-entropy loss of larger models is overwhelmingly driven by frequent word losses mirroring the effect of increased vocabulary size. However, unlike the pattern in figure \ref{fig:3a}, scaling up model size does not exacerbate errors on infrequent tokens. Figure \ref{fig:5b} demonstrates that the global cross-entropy loss declines as model size increases, showing the same tendency of scaling up the vocabulary size (figure \ref{fig:3b}).} 
  \label{fig:model size}
\end{figure}

\subsection{Parameter scaling recovers the same frequent-word gains} \label{main:model size}

Thus far, we have shown that enlarging the vocabulary size decreases word-level average per-word loss on frequent words, which translates into lower cross‑entropy loss and better downstream task performance. One might ask whether similar gains could be achieved without altering the vocabulary size by adjusting other model hyperparameters. Our experiment illustrates that increasing the model’s parameter count can replicate the same benefit. To investigate this, we use the Pythia suite, where each checkpoint is trained on the same dataset with identical hyperparameters, differing only in model size \cite{biderman2023pythiasuiteanalyzinglarge}. We also examine the OLMo-2 series \cite{olmo2} to determine whether the same pattern persists in contemporary large-scale language models (see the Appendix \ref{apdx:olmo}). 

Figure \ref{fig:model size} compares Pythia models ($160$M, $1$B, $6.9$B parameter count) in terms of word-level average per-word loss, proportion of frequent words losses on total loss, and global cross-entropy loss measured on the Paloma validation dataset \cite{magnusson2024palomabenchmarkevaluatinglanguage}. Figure \ref{fig:5a} shows that larger models predict frequent words far more accurately: for the most frequent $2,500$ words in FineWeb-Edu, the average per‑word loss drops from $4.48$ nats ($160$M) to $3.76$ nats ($1$B) and $3.38$ nats ($6.9$B). Furthermore, frequent words loss dominates the total loss as model size grows, which accounts for almost $70$\% of the total loss in the Pythia $1$B and $6.9$B model size, compared with roughly $48$\% in the Pythia $160$M. Unlike the trend observed when only the vocabulary size is increased (Figure \ref{fig:3b}), scaling model capacity does not inflate loss on rare tokens; their share of the loss shrinks. Figure \ref{fig:5b} illustrates that the global cross-entropy loss on the Paloma validation set drops from $4.32$ nats in the Pythia $160$M to $2.51$ nats and $2.26$ nats in the Pythia $1$B and $6.9$B respectively, mirroring the loss reduction seen when expanding the vocabulary (Figure \ref{fig:scaling vocab} and \ref{fig:scaling 30B}). From this experiment, we can observe that enlarging the model size also lowers loss on frequent words while effectively ignoring rare token losses, so the global cross‑entropy reduction is larger than what scaling up vocabulary size alone can deliver. 



\section{Discussion}

\begin{wraptable}[17]{r}{0.45\columnwidth}  
\vspace{-1.0\baselineskip}
  \centering
  \caption{Upper bound of Kolmogorov complexity ($K(X^{N})$) and NCR for the $45.97$ billion‑byte FineWeb‑Edu corpus tokenized with various SuperBPE variants. $t$ and Avg denote the stage-switch vocabulary threshold and an average downstream performance reported by \citep{superbpe}, respectively. The SuperBPE variant with the highest average performance exhibits the lowest tokenized text complexity. }
  \label{tab:superbpe}
  \resizebox{\linewidth}{!}{
    \begin{tabular}{lccr}
      \toprule
      \textbf{superBPE} & \textbf{$K(X^{N})$} & \textbf{NCR} & \textbf{Avg \citep{superbpe}} \\
      \midrule[1.5pt]
      \multirow{1}{*}{\textsc{$200\mathrm{K} (BPE)$}}   & $10.21B$     & 0.222  & 39.8 \\
      \midrule
      \multirow{1}{*}{\textsc{$200\mathrm{K} (t=180K)$}}  & $10.03B$     & 0.218  & 43.8 \\
      \midrule
      \multirow{1}{*}{\textsc{$200\mathrm{K} (t=160K)$}}    & $10.05B$     & 0.219  & 43.4 \\
      \midrule
      \multirow{1}{*}{\textsc{$200\mathrm{K} (t=80K)$}}    & $10.10B$     & 0.220  & 42.9  \\
      \bottomrule
    \end{tabular}%
  }
\end{wraptable}

\subsection{Can we reduce the tokenized text complexity without intensifying frequency imbalance?}

A larger vocabulary reduces the tokenized text complexity by decreasing the number of tokens and sharpening the token-frequency imbalance. However, Section \ref{main:token freq} shows that a sharper token-frequency imbalance increases rare-token loss. This raises the question: can we lower tokenized text complexity without enlarging the vocabulary size? SuperBPE \citep{superbpe} is one such example.

SuperBPE employs a two‑stage BPE algorithm: in the first stage (up to a threshold vocabulary size $t$), it operates identically to standard BPE, while in the second stage, it abandons whitespace pre‑tokenization. By permitting merges across whitespace boundaries, Superbpe directs subsequent merges toward frequent tokens, thereby limiting the introduction of new rare tokens, and preventing further token frequency imbalance as the vocabulary expands. 

Table \ref{tab:superbpe} reports the upper bound Kolmogorov complexity, NCR of the $45.97$ billion‑byte FineWeb‑Edu corpus tokenized with various SuperBPE variants, along with average downstream performance reported by \citep{superbpe}. The result shows that the superBPE achieves a lower token count and a less skewed token-frequency distribution than BPE at equal vocabulary. Also, the superBPE variant with the highest average performance exhibits the lowest tokenized text complexity. This finding demonstrates that both vocabulary expansion and the superBPE variants leverage the same benefit: they reduce the complexity of the tokenized text by segmenting common character sequences into single tokens, thereby reducing the number of statistical patterns that models need to learn.

\subsection{Deduplication through the lens of vocabulary frequency imbalance}

Deduplication removes exact or near-duplicate content from a corpus, reducing repetition and leakage so that a larger fraction of the dataset is novel \cite{Lee2022,penedo2024finewebdatasetsdecantingweb,Li2024}. Deduplication can increase the information density of a dataset, i.e., the fraction of non-redundant data in the corpus. For non-i.i.d.\ data, information density can be proxied by the byte-level entropy rate, the theoretical minimum amount of non-redundant information. However, the entropy rate is only tightly approximable with near infinite context compressors, finite window compressors can distort comparisons before and after deduplication. A more practical loose upper bound is the byte-level bits-per-byte (BPB): under an i.i.d.\ assumption it quantifies information density, with unigram (byte) entropy upper-bounding the true entropy rate. Under the byte-level vocabulary, greater information density implies a higher BPB, driving entropy toward \(\log_2 256 = 8\) bits/byte, which corresponds to a more uniform distribution and lower frequency imbalance. Exploring how deduplication alters BPB, information density, and vocabulary frequency imbalance is a promising avenue for future work.

\subsection{Why higher token‑frequency imbalance degrades machine translation task performance?}

\citet{zouhar_tokenization} reported that higher token-frequency imbalance degrades machine translation task performance, contrary to our findings. What explains this contrasting behavior between machine translation and monolingual settings? We presume that this is related to rare word issues in machine translation tasks \cite{senrich-rareword,senrich2019,koehn2017-sixchallenge,luong2015,gu-token-level-adaptive}.




In machine translation, the source and target vocabularies overlap minimally: vocabularies that are common in English often occur rarely in French, and vice versa. Conversely, in a monolingual setting, the pre-training corpus and downstream benchmarks share not only vocabularies but also many of the same frequent words, resulting in extensive text overlap. Consequently, BLEU penalizes every missing target-side n-gram, so mistranslating even a handful of infrequent tokens can sharply lower scores. To address this rare-word challenge, methods like vocabulary trimming deliberately shrink the vocabulary size to reduce the impact of low-frequency tokens on translation performance \cite{gowda2020,bpe_dropout,bpegetspicky}.

\section{Related Work}\label{sec:relatedwork}
\paragraph{Impact of tokenization on transformer language models}
Empirical and theoretical evidence shows that tokenization is a central determinant of both the quality and the speed with which Transformers learn. Early byte or character-level systems such as CANINE \cite{canine} and ByT5 \cite{byt5} avoid sub‑word vocabularies but pay a steep price: sequences become an order of magnitude longer, gradients are noisier, and convergence is markedly slower than for sub‑word models. Enlarging the BPE dictionary lets even an unigram model approach near-optimal cross-entropy by approximating word-level tokens, whereas the same model without a tokenizer underfits \citep{rajaraman2025theorytokenizationllms}. Furthermore, Controlled scaling studies reveal a log–linear pattern: exponentially expanding the input vocabulary leads to an almost linear drop in loss across model sizes \cite{tao2024scalinglawsvocabularylarger}. \citet{huang2025overtokenizedtransformervocabularygenerally} push this further with “Over‑Tokenized Transformers,” showing that a $400$M parameter model with a $12.8$M token encoder matches the loss of a $1$B model baseline without extra compute. 

Beyond the gains from larger vocabularies, a growing body of work shows that permitting merges across word boundaries (i.e., disabling whitespace pretokenization) improves language model performance. \citet{superbpe,schmidt2025boundlessbytepairencoding} turn off white-space pretokenization during BPE tokenizer training once the vocabulary reaches $t$, or at frequency-driven transition points. Taken together, these results indicate that without tokenization, language models stall at character granularity, but with a tokenizer, they converge faster and generalize better.



\paragraph{Large language model performs a two stage lossless compression}
Large language models perform a two stage lossless compression where tokenizer acts as a pre-compressor and transformer models the tokenized data \cite{deletang_modeling,Lester2024}. \citet{Lester2024} further explore the role of tokenizer as a pre-compressor by replacing it with small transformer model with arithmetic coding, and models trained on neurally compressed text underperforms subword tokenizers but beat byte-level LMs. \citet{deletang_modeling} and \citet{llmzip} has been proposed that transformer language models can be harnessed for compression via arithmetic coding and \citet{Depeiges2025} extends this line by probing the possibility of transformers as a universal compressor across text, images, and audio. However, most literatures mixuse the concept of generalization and compression where generalization focuses on predicting unseen data from different datasets by exploiting shared statistical patterns while compression focuses on encoding a fixed dataset in as few bits as possible by modeling its distribution and removing redundancy. This work investigates the role and impact of tokenizer vocabulary size on generalization through the lens of data compression and information theory. 

\paragraph{BPE tokenizer 
with inductive biases}
BPE tokenizers typically adopt pre-tokenization where regular expressions split text into chunks, sometimes called pre-tokens. Recent work shows that removing pre-tokenization degrades downstream performance across diverse tokenizers \cite{dagan2024gettingtokenizerpretrainingdomain, schmidt2024}. Pre-tokenization rules act as an inductive bias grounded in morphology and grammar, helping the tokenizer capture ubiquitous statistical patterns in natural text, which is crucial for transformer language models to generalize unseen data.  


\section{Conclusion}
This work set out to explain \emph{how} and \emph{why} larger vocabularies boost language model performance. Our experiments reveal a single, robust mechanism: enlarging the vocabulary reduces tokenized text complexity by segmenting frequent non-i.i.d. patterns into single tokens which in turn decreases the number of statistical patterns the language models learn during pre-training, and ultimately lowering language modeling difficulty. Once a vocabulary reaches roughly $24$K size, every common word is already a single token; subsequent growth therefore minimally refine segmentation but instead steepens the long-tailed frequency distribution, focusing optimization on the same frequent words and driving loss down. Heavy overlap in common words explains downstream performance transfer, while enlarging model parameters with a fixed tokenizer reproduces the benefit, pointing to a shared optimization dynamic between vocabulary and model scaling. Recognising this fact turns Kolmogorov complexity of data into a principled dial for tokenizer–model co-design and sharpens our understanding of the scaling forces that govern language-model pre-training. We therefore conclude:

\begin{quote}
\centering
\emph{Expanding the tokenizer mainly reduces uncertainty for the most common words, with little payoff for the rare tail.}
\end{quote}


\newpage
\section*{Acknowledgement}
This work was supported by Institute for Information \& communications Technology Planning \& Evaluation (IITP) grant funded by the Korea government(MSIT) (RS-2019-II190075, Artificial Intelligence Graduate School Program(KAIST)), and Institute of Information \& communications Technology Planning \& Evaluation (IITP) grant funded by the Korea government(MSIT) (No.RS-2022-II220184, 2022-0-00184, Development and Study of AI Technologies to Inexpensively Conform to Evolving Policy on Ethics).

\bibliographystyle{abbrvnat}
\bibliography{cite.bib}

@inproceedings{rajaraman2025theorytokenizationllms,
  author       = {Nived Rajaraman and
                  Jiantao Jiao and
                  Kannan Ramchandran},
  editor       = {Amir Globersons and
                  Lester Mackey and
                  Danielle Belgrave and
                  Angela Fan and
                  Ulrich Paquet and
                  Jakub M. Tomczak and
                  Cheng Zhang},
  title        = {An Analysis of Tokenization: Transformers under Markov Data},
  booktitle    = {Advances in Neural Information Processing Systems 38: Annual Conference
                  on Neural Information Processing Systems 2024, NeurIPS 2024, Vancouver,
                  BC, Canada, December 10 - 15, 2024},
  year         = {2024},
  url          = {http://papers.nips.cc/paper\_files/paper/2024/hash/724afcaae4ae92a9220a077ffe80088d-Abstract-Conference.html},
  bibsource    = {dblp computer science bibliography, https://dblp.org}
}

@ARTICLE{englishshannonentropy,
  author={Shannon, C. E.},
  journal={The Bell System Technical Journal}, 
  title={Prediction and entropy of printed English}, 
  year={1951},
  volume={30},
  number={1},
  pages={50-64},
  doi={10.1002/j.1538-7305.1951.tb01366.x}}

@inproceedings{magnusson2024palomabenchmarkevaluatinglanguage,
  author       = {Ian Magnusson and
                  Akshita Bhagia and
                  Valentin Hofmann and
                  Luca Soldaini and
                  Ananya Harsh Jha and
                  Oyvind Tafjord and
                  Dustin Schwenk and
                  Evan Pete Walsh and
                  Yanai Elazar and
                  Kyle Lo and
                  Dirk Groeneveld and
                  Iz Beltagy and
                  Hanna Hajishirzi and
                  Noah A. Smith and
                  Kyle Richardson and
                  Jesse Dodge},
  editor       = {Amir Globersons and
                  Lester Mackey and
                  Danielle Belgrave and
                  Angela Fan and
                  Ulrich Paquet and
                  Jakub M. Tomczak and
                  Cheng Zhang},
  title        = {Paloma: {A} Benchmark for Evaluating Language Model Fit},
  booktitle    = {Advances in Neural Information Processing Systems 38: Annual Conference
                  on Neural Information Processing Systems 2024, NeurIPS 2024, Vancouver,
                  BC, Canada, December 10 - 15, 2024},
  year         = {2024},
  url          = {http://papers.nips.cc/paper\_files/paper/2024/hash/760b2d94398aa61468aa3bc11506d9ea-Abstract-Datasets\_and\_Benchmarks\_Track.html},
  bibsource    = {dblp computer science bibliography, https://dblp.org}
}

@article{Jelinek1977Perplexity,
  title={Perplexity—a measure of the difficulty of speech recognition tasks},
  author={Frederick Jelinek and Robert L. Mercer and Lalit R. Bahl and Janet M. Baker},
  journal={Journal of the Acoustical Society of America},
  year={1977},
  volume={62},
  url={https://api.semanticscholar.org/CorpusID:121680873}
}

@inproceedings{dagan2024gettingtokenizerpretrainingdomain,
  author       = {Gautier Dagan and
                  Gabriel Synnaeve and
                  Baptiste Rozi{\`{e}}re},
  title        = {Getting the most out of your tokenizer for pre-training and domain
                  adaptation},
  booktitle    = {Forty-first International Conference on Machine Learning, {ICML} 2024,
                  Vienna, Austria, July 21-27, 2024},
  publisher    = {OpenReview.net},
  year         = {2024},
  url          = {https://openreview.net/forum?id=ZFYBnLljtT},
  bibsource    = {dblp computer science bibliography, https://dblp.org}
}

@article{huang2025overtokenizedtransformervocabularygenerally,
  author       = {Hongzhi Huang and
                  Defa Zhu and
                  Banggu Wu and
                  Yutao Zeng and
                  Ya Wang and
                  Qiyang Min and
                  Xun Zhou},
  title        = {Over-Tokenized Transformer: Vocabulary is Generally Worth Scaling},
  journal      = {CoRR},
  volume       = {abs/2501.16975},
  year         = {2025},
  url          = {https://doi.org/10.48550/arXiv.2501.16975},
  doi          = {10.48550/ARXIV.2501.16975},
  eprinttype    = {arXiv},
  eprint       = {2501.16975},
  bibsource    = {dblp computer science bibliography, https://dblp.org}
}

@inproceedings{tao2024scalinglawsvocabularylarger,
  author       = {Chaofan Tao and
                  Qian Liu and
                  Longxu Dou and
                  Niklas Muennighoff and
                  Zhongwei Wan and
                  Ping Luo and
                  Min Lin and
                  Ngai Wong},
  editor       = {Amir Globersons and
                  Lester Mackey and
                  Danielle Belgrave and
                  Angela Fan and
                  Ulrich Paquet and
                  Jakub M. Tomczak and
                  Cheng Zhang},
  title        = {Scaling Laws with Vocabulary: Larger Models Deserve Larger Vocabularies},
  booktitle    = {Advances in Neural Information Processing Systems 38: Annual Conference
                  on Neural Information Processing Systems 2024, NeurIPS 2024, Vancouver,
                  BC, Canada, December 10 - 15, 2024},
  year         = {2024},
  url          = {http://papers.nips.cc/paper\_files/paper/2024/hash/cf5a019ae9c11b4be88213ce3f85d85c-Abstract-Conference.html},
  bibsource    = {dblp computer science bibliography, https://dblp.org}
}

@inproceedings{schmidt2024tokenizationcompression,
  author       = {Craig W. Schmidt and
                  Varshini Reddy and
                  Haoran Zhang and
                  Alec Alameddine and
                  Omri Uzan and
                  Yuval Pinter and
                  Chris Tanner},
  editor       = {Yaser Al{-}Onaizan and
                  Mohit Bansal and
                  Yun{-}Nung Chen},
  title        = {Tokenization Is More Than Compression},
  booktitle    = {Proceedings of the 2024 Conference on Empirical Methods in Natural
                  Language Processing, {EMNLP} 2024, Miami, FL, USA, November 12-16,
                  2024},
  pages        = {678--702},
  publisher    = {Association for Computational Linguistics},
  year         = {2024},
  url          = {https://aclanthology.org/2024.emnlp-main.40},
  bibsource    = {dblp computer science bibliography, https://dblp.org}
}

@inproceedings{ali2024tokenizerchoicellmtraining,
  author       = {Mehdi Ali and
                  Michael Fromm and
                  Klaudia Thellmann and
                  Richard Rutmann and
                  Max L{\"{u}}bbering and
                  Johannes Leveling and
                  Katrin Klug and
                  Jan Ebert and
                  Niclas Doll and
                  Jasper Schulze Buschhoff and
                  Charvi Jain and
                  Alexander Arno Weber and
                  Lena Jurkschat and
                  Hammam Abdelwahab and
                  Chelsea John and
                  Pedro Ortiz Suarez and
                  Malte Ostendorff and
                  Samuel Weinbach and
                  Rafet Sifa and
                  Stefan Kesselheim and
                  Nicolas Flores{-}Herr},
  editor       = {Kevin Duh and
                  Helena G{\'{o}}mez{-}Adorno and
                  Steven Bethard},
  title        = {Tokenizer Choice For {LLM} Training: Negligible or Crucial?},
  booktitle    = {Findings of the Association for Computational Linguistics: {NAACL}
                  2024, Mexico City, Mexico, June 16-21, 2024},
  pages        = {3907--3924},
  publisher    = {Association for Computational Linguistics},
  year         = {2024},
  url          = {https://doi.org/10.18653/v1/2024.findings-naacl.247},
  doi          = {10.18653/V1/2024.FINDINGS-NAACL.247},
  bibsource    = {dblp computer science bibliography, https://dblp.org}
}

@article{reddy2025enoughdiminishingreturnstokenization,
  author       = {Varshini Reddy and
                  Craig W. Schmidt and
                  Yuval Pinter and
                  Chris Tanner},
  title        = {How Much is Enough? The Diminishing Returns of Tokenization Training
                  Data},
  journal      = {CoRR},
  volume       = {abs/2502.20273},
  year         = {2025},
  url          = {https://doi.org/10.48550/arXiv.2502.20273},
  doi          = {10.48550/ARXIV.2502.20273},
  eprinttype    = {arXiv},
  eprint       = {2502.20273},
  bibsource    = {dblp computer science bibliography, https://dblp.org}
}

@article{huang2024compressionrepresentsintelligencelinearly,
  author       = {Yuzhen Huang and
                  Jinghan Zhang and
                  Zifei Shan and
                  Junxian He},
  title        = {Compression Represents Intelligence Linearly},
  journal      = {CoRR},
  volume       = {abs/2404.09937},
  year         = {2024},
  url          = {https://doi.org/10.48550/arXiv.2404.09937},
  doi          = {10.48550/ARXIV.2404.09937},
  eprinttype    = {arXiv},
  eprint       = {2404.09937},
  bibsource    = {dblp computer science bibliography, https://dblp.org}
}

@inproceedings{penedo2024finewebdatasetsdecantingweb,
  author       = {Guilherme Penedo and
                  Hynek Kydl{\'{\i}}cek and
                  Loubna Ben Allal and
                  Anton Lozhkov and
                  Margaret Mitchell and
                  Colin A. Raffel and
                  Leandro von Werra and
                  Thomas Wolf},
  editor       = {Amir Globersons and
                  Lester Mackey and
                  Danielle Belgrave and
                  Angela Fan and
                  Ulrich Paquet and
                  Jakub M. Tomczak and
                  Cheng Zhang},
  title        = {The FineWeb Datasets: Decanting the Web for the Finest Text Data at
                  Scale},
  booktitle    = {Advances in Neural Information Processing Systems 38: Annual Conference
                  on Neural Information Processing Systems 2024, NeurIPS 2024, Vancouver,
                  BC, Canada, December 10 - 15, 2024},
  year         = {2024},
  url          = {http://papers.nips.cc/paper\_files/paper/2024/hash/370df50ccfdf8bde18f8f9c2d9151bda-Abstract-Datasets\_and\_Benchmarks\_Track.html},
  bibsource    = {dblp computer science bibliography, https://dblp.org}
}

@misc{Gokaslan2019OpenWeb,
    title={OpenWebText Corpus},
    author={Gokaslan, Aaron and Cohen, Vanya and Pavlick, Ellie and Tellex, Stefanie},
    howpublished={\url{http://Skylion007.github.io/OpenWebTextCorpus}},
    year={2019}
}

@article{clark2018thinksolvedquestionanswering,
  author       = {Sumithra Bhakthavatsalam and
                  Daniel Khashabi and
                  Tushar Khot and
                  Bhavana Dalvi Mishra and
                  Kyle Richardson and
                  Ashish Sabharwal and
                  Carissa Schoenick and
                  Oyvind Tafjord and
                  Peter Clark},
  title        = {Think you have Solved Direct-Answer Question Answering? Try ARC-DA,
                  the Direct-Answer {AI2} Reasoning Challenge},
  journal      = {CoRR},
  volume       = {abs/2102.03315},
  year         = {2021},
  url          = {https://arxiv.org/abs/2102.03315},
  eprinttype    = {arXiv},
  eprint       = {2102.03315},
  bibsource    = {dblp computer science bibliography, https://dblp.org}
}

@inproceedings{zellers2019hellaswagmachinereallyfinish,
  author       = {Rowan Zellers and
                  Ari Holtzman and
                  Yonatan Bisk and
                  Ali Farhadi and
                  Yejin Choi},
  editor       = {Anna Korhonen and
                  David R. Traum and
                  Llu{\'{\i}}s M{\`{a}}rquez},
  title        = {HellaSwag: Can a Machine Really Finish Your Sentence?},
  booktitle    = {Proceedings of the 57th Conference of the Association for Computational
                  Linguistics, {ACL} 2019, Florence, Italy, July 28- August 2, 2019,
                  Volume 1: Long Papers},
  pages        = {4791--4800},
  publisher    = {Association for Computational Linguistics},
  year         = {2019},
  url          = {https://doi.org/10.18653/v1/p19-1472},
  doi          = {10.18653/V1/P19-1472},
  bibsource    = {dblp computer science bibliography, https://dblp.org}
}

@inproceedings{sciq,
  author       = {Johannes Welbl and
                  Nelson F. Liu and
                  Matt Gardner},
  editor       = {Leon Derczynski and
                  Wei Xu and
                  Alan Ritter and
                  Tim Baldwin},
  title        = {Crowdsourcing Multiple Choice Science Questions},
  booktitle    = {Proceedings of the 3rd Workshop on Noisy User-generated Text, NUT@EMNLP
                  2017, Copenhagen, Denmark, September 7, 2017},
  pages        = {94--106},
  publisher    = {Association for Computational Linguistics},
  year         = {2017},
  url          = {https://doi.org/10.18653/v1/w17-4413},
  doi          = {10.18653/V1/W17-4413},
  bibsource    = {dblp computer science bibliography, https://dblp.org}
}

@inproceedings{bisk2019piqareasoningphysicalcommonsense,
  author       = {Yonatan Bisk and
                  Rowan Zellers and
                  Ronan Le Bras and
                  Jianfeng Gao and
                  Yejin Choi},
  title        = {{PIQA:} Reasoning about Physical Commonsense in Natural Language},
  booktitle    = {The Thirty-Fourth {AAAI} Conference on Artificial Intelligence, {AAAI}
                  2020, The Thirty-Second Innovative Applications of Artificial Intelligence
                  Conference, {IAAI} 2020, The Tenth {AAAI} Symposium on Educational
                  Advances in Artificial Intelligence, {EAAI} 2020, New York, NY, USA,
                  February 7-12, 2020},
  pages        = {7432--7439},
  publisher    = {{AAAI} Press},
  year         = {2020},
  url          = {https://doi.org/10.1609/aaai.v34i05.6239},
  doi          = {10.1609/AAAI.V34I05.6239},
  bibsource    = {dblp computer science bibliography, https://dblp.org}
}

@inproceedings{biderman2023pythiasuiteanalyzinglarge,
  author       = {Stella Biderman and
                  Hailey Schoelkopf and
                  Quentin Gregory Anthony and
                  Herbie Bradley and
                  Kyle O'Brien and
                  Eric Hallahan and
                  Mohammad Aflah Khan and
                  Shivanshu Purohit and
                  USVSN Sai Prashanth and
                  Edward Raff and
                  Aviya Skowron and
                  Lintang Sutawika and
                  Oskar van der Wal},
  editor       = {Andreas Krause and
                  Emma Brunskill and
                  Kyunghyun Cho and
                  Barbara Engelhardt and
                  Sivan Sabato and
                  Jonathan Scarlett},
  title        = {Pythia: {A} Suite for Analyzing Large Language Models Across Training
                  and Scaling},
  booktitle    = {International Conference on Machine Learning, {ICML} 2023, 23-29 July
                  2023, Honolulu, Hawaii, {USA}},
  series       = {Proceedings of Machine Learning Research},
  volume       = {202},
  pages        = {2397--2430},
  publisher    = {{PMLR}},
  year         = {2023},
  url          = {https://proceedings.mlr.press/v202/biderman23a.html},
  bibsource    = {dblp computer science bibliography, https://dblp.org}
}

@article{pinto2024fairlanguagemodelparadox,
  author       = {Andrea Pinto and
                  Tomer Galanti and
                  Randall Balestriero},
  title        = {The Fair Language Model Paradox},
  journal      = {CoRR},
  volume       = {abs/2410.11985},
  year         = {2024},
  url          = {https://doi.org/10.48550/arXiv.2410.11985},
  doi          = {10.48550/ARXIV.2410.11985},
  eprinttype    = {arXiv},
  eprint       = {2410.11985},
  bibsource    = {dblp computer science bibliography, https://dblp.org}
}

@article{kim2025perilnrevisitinglayernormalization,
  author       = {Jeonghoon Kim and
                  Byeongchan Lee and
                  Cheonbok Park and
                  Yeontaek Oh and
                  Beomjun Kim and
                  Taehwan Yoo and
                  Seongjin Shin and
                  Dongyoon Han and
                  Jinwoo Shin and
                  Kang Min Yoo},
  title        = {Peri-LN: Revisiting Layer Normalization in the Transformer Architecture},
  journal      = {CoRR},
  volume       = {abs/2502.02732},
  year         = {2025},
  url          = {https://doi.org/10.48550/arXiv.2502.02732},
  doi          = {10.48550/ARXIV.2502.02732},
  eprinttype    = {arXiv},
  eprint       = {2502.02732},
  bibsource    = {dblp computer science bibliography, https://dblp.org}
}

@inproceedings{
mircea2024gradient,
title={Gradient Dissent in Language Model Training and Saturation},
author={Andrei Mircea and Ekaterina Lobacheva and Irina Rish},
booktitle={High-dimensional Learning Dynamics 2024: The Emergence of Structure and Reasoning},
year={2024},
url={https://openreview.net/forum?id=tJj3psv9nm}
}

@inproceedings{xiong2020layernormalizationtransformerarchitecture,
  author       = {Ruibin Xiong and
                  Yunchang Yang and
                  Di He and
                  Kai Zheng and
                  Shuxin Zheng and
                  Chen Xing and
                  Huishuai Zhang and
                  Yanyan Lan and
                  Liwei Wang and
                  Tie{-}Yan Liu},
  title        = {On Layer Normalization in the Transformer Architecture},
  booktitle    = {Proceedings of the 37th International Conference on Machine Learning,
                  {ICML} 2020, 13-18 July 2020, Virtual Event},
  series       = {Proceedings of Machine Learning Research},
  volume       = {119},
  pages        = {10524--10533},
  publisher    = {{PMLR}},
  year         = {2020},
  url          = {http://proceedings.mlr.press/v119/xiong20b.html},
  bibsource    = {dblp computer science bibliography, https://dblp.org}
}

@inproceedings{hayase2024data,
  author       = {Jonathan Hayase and
                  Alisa Liu and
                  Yejin Choi and
                  Sewoong Oh and
                  Noah A. Smith},
  editor       = {Amir Globersons and
                  Lester Mackey and
                  Danielle Belgrave and
                  Angela Fan and
                  Ulrich Paquet and
                  Jakub M. Tomczak and
                  Cheng Zhang},
  title        = {Data Mixture Inference Attack: {BPE} Tokenizers Reveal Training Data
                  Compositions},
  booktitle    = {Advances in Neural Information Processing Systems 38: Annual Conference
                  on Neural Information Processing Systems 2024, NeurIPS 2024, Vancouver,
                  BC, Canada, December 10 - 15, 2024},
  year         = {2024},
  url          = {http://papers.nips.cc/paper\_files/paper/2024/hash/10e6dfea9a673bef4a7b1cb9234891bc-Abstract-Conference.html},
  bibsource    = {dblp computer science bibliography, https://dblp.org}
}

@inproceedings{zouhar_tokenization,
  author       = {Vil{\'{e}}m Zouhar and
                  Clara Meister and
                  Juan Luis Gastaldi and
                  Li Du and
                  Mrinmaya Sachan and
                  Ryan Cotterell},
  editor       = {Anna Rogers and
                  Jordan L. Boyd{-}Graber and
                  Naoaki Okazaki},
  title        = {Tokenization and the Noiseless Channel},
  booktitle    = {Proceedings of the 61st Annual Meeting of the Association for Computational
                  Linguistics (Volume 1: Long Papers), {ACL} 2023, Toronto, Canada,
                  July 9-14, 2023},
  pages        = {5184--5207},
  publisher    = {Association for Computational Linguistics},
  year         = {2023},
  url          = {https://doi.org/10.18653/v1/2023.acl-long.284},
  doi          = {10.18653/V1/2023.ACL-LONG.284},
  bibsource    = {dblp computer science bibliography, https://dblp.org}
}

@inproceedings{senrich-rareword,
  author       = {Rico Sennrich and
                  Barry Haddow and
                  Alexandra Birch},
  title        = {Neural Machine Translation of Rare Words with Subword Units},
  booktitle    = {Proceedings of the 54th Annual Meeting of the Association for Computational
                  Linguistics, {ACL} 2016, August 7-12, 2016, Berlin, Germany, Volume
                  1: Long Papers},
  publisher    = {The Association for Computer Linguistics},
  year         = {2016},
  url          = {https://doi.org/10.18653/v1/p16-1162},
  doi          = {10.18653/V1/P16-1162},
  bibsource    = {dblp computer science bibliography, https://dblp.org}
}

@inproceedings{gu-token-level-adaptive,
  author       = {Shuhao Gu and
                  Jinchao Zhang and
                  Fandong Meng and
                  Yang Feng and
                  Wanying Xie and
                  Jie Zhou and
                  Dong Yu},
  editor       = {Bonnie Webber and
                  Trevor Cohn and
                  Yulan He and
                  Yang Liu},
  title        = {Token-level Adaptive Training for Neural Machine Translation},
  booktitle    = {Proceedings of the 2020 Conference on Empirical Methods in Natural
                  Language Processing, {EMNLP} 2020, Online, November 16-20, 2020},
  pages        = {1035--1046},
  publisher    = {Association for Computational Linguistics},
  year         = {2020},
  url          = {https://doi.org/10.18653/v1/2020.emnlp-main.76},
  doi          = {10.18653/V1/2020.EMNLP-MAIN.76},
  bibsource    = {dblp computer science bibliography, https://dblp.org}
}

@inproceedings{senrich2019,
  author       = {Rico Sennrich and
                  Biao Zhang},
  editor       = {Anna Korhonen and
                  David R. Traum and
                  Llu{\'{\i}}s M{\`{a}}rquez},
  title        = {Revisiting Low-Resource Neural Machine Translation: {A} Case Study},
  booktitle    = {Proceedings of the 57th Conference of the Association for Computational
                  Linguistics, {ACL} 2019, Florence, Italy, July 28- August 2, 2019,
                  Volume 1: Long Papers},
  pages        = {211--221},
  publisher    = {Association for Computational Linguistics},
  year         = {2019},
  url          = {https://doi.org/10.18653/v1/p19-1021},
  doi          = {10.18653/V1/P19-1021},
  bibsource    = {dblp computer science bibliography, https://dblp.org}
}

@inproceedings{luong2015,
  author       = {Thang Luong and
                  Ilya Sutskever and
                  Quoc V. Le and
                  Oriol Vinyals and
                  Wojciech Zaremba},
  title        = {Addressing the Rare Word Problem in Neural Machine Translation},
  booktitle    = {Proceedings of the 53rd Annual Meeting of the Association for Computational
                  Linguistics and the 7th International Joint Conference on Natural
                  Language Processing of the Asian Federation of Natural Language Processing,
                  {ACL} 2015, July 26-31, 2015, Beijing, China, Volume 1: Long Papers},
  pages        = {11--19},
  publisher    = {The Association for Computer Linguistics},
  year         = {2015},
  url          = {https://doi.org/10.3115/v1/p15-1002},
  doi          = {10.3115/V1/P15-1002},
  bibsource    = {dblp computer science bibliography, https://dblp.org}
}

@inproceedings{koehn2017-sixchallenge,
  author       = {Philipp Koehn and
                  Rebecca Knowles},
  editor       = {Thang Luong and
                  Alexandra Birch and
                  Graham Neubig and
                  Andrew M. Finch},
  title        = {Six Challenges for Neural Machine Translation},
  booktitle    = {Proceedings of the First Workshop on Neural Machine Translation, NMT@ACL
                  2017, Vancouver, Canada, August 4, 2017},
  pages        = {28--39},
  publisher    = {Association for Computational Linguistics},
  year         = {2017},
  url          = {https://doi.org/10.18653/v1/w17-3204},
  doi          = {10.18653/V1/W17-3204},
  bibsource    = {dblp computer science bibliography, https://dblp.org}
}

@inproceedings{gowda2020,
  author       = {Thamme Gowda and
                  Jonathan May},
  editor       = {Trevor Cohn and
                  Yulan He and
                  Yang Liu},
  title        = {Finding the Optimal Vocabulary Size for Neural Machine Translation},
  booktitle    = {Findings of the Association for Computational Linguistics: {EMNLP}
                  2020, Online Event, 16-20 November 2020},
  series       = {Findings of {ACL}},
  volume       = {{EMNLP} 2020},
  pages        = {3955--3964},
  publisher    = {Association for Computational Linguistics},
  year         = {2020},
  url          = {https://doi.org/10.18653/v1/2020.findings-emnlp.352},
  doi          = {10.18653/V1/2020.FINDINGS-EMNLP.352},
  bibsource    = {dblp computer science bibliography, https://dblp.org}
}

@inproceedings{bpe_dropout,
  author       = {Ivan Provilkov and
                  Dmitrii Emelianenko and
                  Elena Voita},
  editor       = {Dan Jurafsky and
                  Joyce Chai and
                  Natalie Schluter and
                  Joel R. Tetreault},
  title        = {BPE-Dropout: Simple and Effective Subword Regularization},
  booktitle    = {Proceedings of the 58th Annual Meeting of the Association for Computational
                  Linguistics, {ACL} 2020, Online, July 5-10, 2020},
  pages        = {1882--1892},
  publisher    = {Association for Computational Linguistics},
  year         = {2020},
  url          = {https://doi.org/10.18653/v1/2020.acl-main.170},
  doi          = {10.18653/V1/2020.ACL-MAIN.170},
  bibsource    = {dblp computer science bibliography, https://dblp.org}
}

@inproceedings{bpegetspicky,
  author       = {Pavel Chizhov and
                  Catherine Arnett and
                  Elizaveta Korotkova and
                  Ivan P. Yamshchikov},
  editor       = {Yaser Al{-}Onaizan and
                  Mohit Bansal and
                  Yun{-}Nung Chen},
  title        = {{BPE} Gets Picky: Efficient Vocabulary Refinement During Tokenizer
                  Training},
  booktitle    = {Proceedings of the 2024 Conference on Empirical Methods in Natural
                  Language Processing, {EMNLP} 2024, Miami, FL, USA, November 12-16,
                  2024},
  pages        = {16587--16604},
  publisher    = {Association for Computational Linguistics},
  year         = {2024},
  url          = {https://aclanthology.org/2024.emnlp-main.925},
  bibsource    = {dblp computer science bibliography, https://dblp.org}
}

@inproceedings{deletang_modeling,
  author       = {Gr{\'{e}}goire Del{\'{e}}tang and
                  Anian Ruoss and
                  Paul{-}Ambroise Duquenne and
                  Elliot Catt and
                  Tim Genewein and
                  Christopher Mattern and
                  Jordi Grau{-}Moya and
                  Li Kevin Wenliang and
                  Matthew Aitchison and
                  Laurent Orseau and
                  Marcus Hutter and
                  Joel Veness},
  title        = {Language Modeling Is Compression},
  booktitle    = {The Twelfth International Conference on Learning Representations,
                  {ICLR} 2024, Vienna, Austria, May 7-11, 2024},
  publisher    = {OpenReview.net},
  year         = {2024},
  url          = {https://openreview.net/forum?id=jznbgiynus},
  bibsource    = {dblp computer science bibliography, https://dblp.org}
}

@article{shannon1948,
  author       = {Claude E. Shannon},
  title        = {A mathematical theory of communication},
  journal      = {Bell Syst. Tech. J.},
  volume       = {27},
  number       = {3},
  pages        = {379--423},
  year         = {1948},
  url          = {https://doi.org/10.1002/j.1538-7305.1948.tb01338.x},
  doi          = {10.1002/J.1538-7305.1948.TB01338.X},
  bibsource    = {dblp computer science bibliography, https://dblp.org}
}

@article{byt5,
  author       = {Linting Xue and
                  Aditya Barua and
                  Noah Constant and
                  Rami Al{-}Rfou and
                  Sharan Narang and
                  Mihir Kale and
                  Adam Roberts and
                  Colin Raffel},
  title        = {ByT5: Towards a Token-Free Future with Pre-trained Byte-to-Byte Models},
  journal      = {Trans. Assoc. Comput. Linguistics},
  volume       = {10},
  pages        = {291--306},
  year         = {2022},
  url          = {https://doi.org/10.1162/tacl\_a\_00461},
  doi          = {10.1162/TACL\_A\_00461},
  bibsource    = {dblp computer science bibliography, https://dblp.org}
}

@article{canine,
  author       = {Jonathan H. Clark and
                  Dan Garrette and
                  Iulia Turc and
                  John Wieting},
  title        = {Canine: Pre-training an Efficient Tokenization-Free Encoder for Language
                  Representation},
  journal      = {Trans. Assoc. Comput. Linguistics},
  volume       = {10},
  pages        = {73--91},
  year         = {2022},
  url          = {https://doi.org/10.1162/tacl\_a\_00448},
  doi          = {10.1162/TACL\_A\_00448},
  bibsource    = {dblp computer science bibliography, https://dblp.org}
}

@article{olmo2,
  author       = {Team OLMo and
                  Pete Walsh and
                  Luca Soldaini and
                  Dirk Groeneveld and
                  Kyle Lo and
                  Shane Arora and
                  Akshita Bhagia and
                  Yuling Gu and
                  Shengyi Huang and
                  Matt Jordan and
                  Nathan Lambert and
                  Dustin Schwenk and
                  Oyvind Tafjord and
                  Taira Anderson and
                  David Atkinson and
                  Faeze Brahman and
                  Christopher Clark and
                  Pradeep Dasigi and
                  Nouha Dziri and
                  Michal Guerquin and
                  Hamish Ivison and
                  Pang Wei Koh and
                  Jiacheng Liu and
                  Saumya Malik and
                  William Merrill and
                  Lester James V. Miranda and
                  Jacob Morrison and
                  Tyler Murray and
                  Crystal Nam and
                  Valentina Pyatkin and
                  Aman Rangapur and
                  Michael Schmitz and
                  Sam Skjonsberg and
                  David Wadden and
                  Christopher Wilhelm and
                  Michael Wilson and
                  Luke Zettlemoyer and
                  Ali Farhadi and
                  Noah A. Smith and
                  Hannaneh Hajishirzi},
  title        = {2 OLMo 2 Furious},
  journal      = {CoRR},
  volume       = {abs/2501.00656},
  year         = {2025},
  url          = {https://doi.org/10.48550/arXiv.2501.00656},
  doi          = {10.48550/ARXIV.2501.00656},
  eprinttype    = {arXiv},
  eprint       = {2501.00656},
  bibsource    = {dblp computer science bibliography, https://dblp.org}
}

@article{gpt-4,
  author       = {OpenAI},
  title        = {{GPT-4} Technical Report},
  journal      = {CoRR},
  volume       = {abs/2303.08774},
  year         = {2023},
  url          = {https://doi.org/10.48550/arXiv.2303.08774},
  doi          = {10.48550/ARXIV.2303.08774},
  eprinttype    = {arXiv},
  eprint       = {2303.08774},
  bibsource    = {dblp computer science bibliography, https://dblp.org}
}

@article{gpt-neox,
  author       = {Sid Black and
                  Stella Biderman and
                  Eric Hallahan and
                  Quentin Anthony and
                  Leo Gao and
                  Laurence Golding and
                  Horace He and
                  Connor Leahy and
                  Kyle McDonell and
                  Jason Phang and
                  Michael Pieler and
                  USVSN Sai Prashanth and
                  Shivanshu Purohit and
                  Laria Reynolds and
                  Jonathan Tow and
                  Ben Wang and
                  Samuel Weinbach},
  title        = {GPT-NeoX-20B: An Open-Source Autoregressive Language Model},
  journal      = {CoRR},
  volume       = {abs/2204.06745},
  year         = {2022},
  url          = {https://doi.org/10.48550/arXiv.2204.06745},
  doi          = {10.48550/ARXIV.2204.06745},
  eprinttype    = {arXiv},
  eprint       = {2204.06745},
  bibsource    = {dblp computer science bibliography, https://dblp.org}
}

@article{yu2025,
  author       = {Da Yu and
                  Edith Cohen and
                  Badih Ghazi and
                  Yangsibo Huang and
                  Pritish Kamath and
                  Ravi Kumar and
                  Daogao Liu and
                  Chiyuan Zhang},
  title        = {Scaling Embedding Layers in Language Models},
  journal      = {CoRR},
  volume       = {abs/2502.01637},
  year         = {2025},
  url          = {https://doi.org/10.48550/arXiv.2502.01637},
  doi          = {10.48550/ARXIV.2502.01637},
  eprinttype    = {arXiv},
  eprint       = {2502.01637},
  bibsource    = {dblp computer science bibliography, https://dblp.org}
}

@inproceedings{benjio2000,
  author       = {Yoshua Bengio and
                  R{\'{e}}jean Ducharme and
                  Pascal Vincent},
  editor       = {Todd K. Leen and
                  Thomas G. Dietterich and
                  Volker Tresp},
  title        = {A Neural Probabilistic Language Model},
  booktitle    = {Advances in Neural Information Processing Systems 13, Papers from
                  Neural Information Processing Systems {(NIPS)} 2000, Denver, CO, {USA}},
  pages        = {932--938},
  publisher    = {{MIT} Press},
  year         = {2000},
  url          = {https://proceedings.neurips.cc/paper/2000/hash/728f206c2a01bf572b5940d7d9a8fa4c-Abstract.html},
  bibsource    = {dblp computer science bibliography, https://dblp.org}
}

@inproceedings{gpt-3,
  author       = {Tom B. Brown and
                  Benjamin Mann and
                  Nick Ryder and
                  Melanie Subbiah and
                  Jared Kaplan and
                  Prafulla Dhariwal and
                  Arvind Neelakantan and
                  Pranav Shyam and
                  Girish Sastry and
                  Amanda Askell and
                  Sandhini Agarwal and
                  Ariel Herbert{-}Voss and
                  Gretchen Krueger and
                  Tom Henighan and
                  Rewon Child and
                  Aditya Ramesh and
                  Daniel M. Ziegler and
                  Jeffrey Wu and
                  Clemens Winter and
                  Christopher Hesse and
                  Mark Chen and
                  Eric Sigler and
                  Mateusz Litwin and
                  Scott Gray and
                  Benjamin Chess and
                  Jack Clark and
                  Christopher Berner and
                  Sam McCandlish and
                  Alec Radford and
                  Ilya Sutskever and
                  Dario Amodei},
  editor       = {Hugo Larochelle and
                  Marc'Aurelio Ranzato and
                  Raia Hadsell and
                  Maria{-}Florina Balcan and
                  Hsuan{-}Tien Lin},
  title        = {Language Models are Few-Shot Learners},
  booktitle    = {Advances in Neural Information Processing Systems 33: Annual Conference
                  on Neural Information Processing Systems 2020, NeurIPS 2020, December
                  6-12, 2020, virtual},
  year         = {2020},
  url          = {https://proceedings.neurips.cc/paper/2020/hash/1457c0d6bfcb4967418bfb8ac142f64a-Abstract.html},
  bibsource    = {dblp computer science bibliography, https://dblp.org}
}

@article{superbpe,
  author       = {Alisa Liu and
                  Jonathan Hayase and
                  Valentin Hofmann and
                  Sewoong Oh and
                  Noah A. Smith and
                  Yejin Choi},
  title        = {SuperBPE: Space Travel for Language Models},
  journal      = {CoRR},
  volume       = {abs/2503.13423},
  year         = {2025},
  url          = {https://doi.org/10.48550/arXiv.2503.13423},
  doi          = {10.48550/ARXIV.2503.13423},
  eprinttype    = {arXiv},
  eprint       = {2503.13423},
  bibsource    = {dblp computer science bibliography, https://dblp.org}
}

@misc{schmidt2025boundlessbytepairencoding,
      title={Boundless Byte Pair Encoding: Breaking the Pre-tokenization Barrier}, 
      author={Craig W. Schmidt and Varshini Reddy and Chris Tanner and Yuval Pinter},
      year={2025},
      eprint={2504.00178},
      archivePrefix={arXiv},
      primaryClass={cs.CL},
      url={https://arxiv.org/abs/2504.00178}, 
}

@article{Reddy2025,
  author       = {Varshini Reddy and
                  Craig W. Schmidt and
                  Yuval Pinter and
                  Chris Tanner},
  title        = {How Much is Enough? The Diminishing Returns of Tokenization Training
                  Data},
  journal      = {CoRR},
  volume       = {abs/2502.20273},
  year         = {2025},
  url          = {https://doi.org/10.48550/arXiv.2502.20273},
  doi          = {10.48550/ARXIV.2502.20273},
  eprinttype    = {arXiv},
  eprint       = {2502.20273},
  bibsource    = {dblp computer science bibliography, https://dblp.org}
}

@inproceedings{BPE,
  author       = {Rico Sennrich and
                  Barry Haddow and
                  Alexandra Birch},
  title        = {Neural Machine Translation of Rare Words with Subword Units},
  booktitle    = {Proceedings of the 54th Annual Meeting of the Association for Computational
                  Linguistics, {ACL} 2016, August 7-12, 2016, Berlin, Germany, Volume
                  1: Long Papers},
  publisher    = {The Association for Computer Linguistics},
  year         = {2016},
  url          = {https://doi.org/10.18653/v1/p16-1162},
  doi          = {10.18653/V1/P16-1162},
  bibsource    = {dblp computer science bibliography, https://dblp.org}
}

@inproceedings{Adamw,
  author       = {Ilya Loshchilov and
                  Frank Hutter},
  title        = {Decoupled Weight Decay Regularization},
  booktitle    = {7th International Conference on Learning Representations, {ICLR} 2019,
                  New Orleans, LA, USA, May 6-9, 2019},
  publisher    = {OpenReview.net},
  year         = {2019},
  url          = {https://openreview.net/forum?id=Bkg6RiCqY7},
  bibsource    = {dblp computer science bibliography, https://dblp.org}
}

@article{goldblum2023,
  author       = {Micah Goldblum and
                  Marc Finzi and
                  Keefer Rowan and
                  Andrew Gordon Wilson},
  title        = {The No Free Lunch Theorem, Kolmogorov Complexity, and the Role of
                  Inductive Biases in Machine Learning},
  journal      = {CoRR},
  volume       = {abs/2304.05366},
  year         = {2023},
  url          = {https://doi.org/10.48550/arXiv.2304.05366},
  doi          = {10.48550/ARXIV.2304.05366},
  eprinttype    = {arXiv},
  eprint       = {2304.05366},
  bibsource    = {dblp computer science bibliography, https://dblp.org}
}

@article{kolmogorov,
  author       = {Andrei N. Kolmogorov},
  title        = {On Tables of Random Numbers (Reprinted from "Sankhya: The Indian Journal
                  of Statistics", Series A, Vol. 25 Part 4, 1963)},
  journal      = {Theor. Comput. Sci.},
  volume       = {207},
  number       = {2},
  pages        = {387--395},
  year         = {1998},
  url          = {https://doi.org/10.1016/S0304-3975(98)00075-9},
  doi          = {10.1016/S0304-3975(98)00075-9},
  bibsource    = {dblp computer science bibliography, https://dblp.org}
}

@article{Morris2025,
  author       = {John X. Morris and
                  Chawin Sitawarin and
                  Chuan Guo and
                  Narine Kokhlikyan and
                  G. Edward Suh and
                  Alexander M. Rush and
                  Kamalika Chaudhuri and
                  Saeed Mahloujifar},
  title        = {How much do language models memorize?},
  journal      = {CoRR},
  volume       = {abs/2505.24832},
  year         = {2025},
  url          = {https://doi.org/10.48550/arXiv.2505.24832},
  doi          = {10.48550/ARXIV.2505.24832},
  eprinttype    = {arXiv},
  eprint       = {2505.24832},
  bibsource    = {dblp computer science bibliography, https://dblp.org}
}

@article{Kolmogorov1968,
  title={Three approaches to the quantitative definition of information},
  author={Andrei N. Kolmogorov},
  journal={International Journal of Computer Mathematics},
  year={1968},
  volume={2},
  pages={157-168},
  url={https://api.semanticscholar.org/CorpusID:119745517}
}

@article{Gage1994,
  title={A new algorithm for data compression},
  author={Philip Gage},
  journal={The C Users Journal archive},
  year={1994},
  volume={12},
  pages={23-38},
  url={https://api.semanticscholar.org/CorpusID:59804030}
}

@article{Lester2024,
  author       = {Brian Lester and
                  Jaehoon Lee and
                  Alexander A. Alemi and
                  Jeffrey Pennington and
                  Adam Roberts and
                  Jascha Sohl{-}Dickstein and
                  Noah Constant},
  title        = {Training LLMs over Neurally Compressed Text},
  journal      = {Trans. Mach. Learn. Res.},
  volume       = {2024},
  year         = {2024},
  url          = {https://openreview.net/forum?id=pRvhMSV48t},
  bibsource    = {dblp computer science bibliography, https://dblp.org}
}

@inproceedings{kolmogorov_test,
  author       = {Ori Yoran and
                  Kunhao Zheng and
                  Fabian Gloeckle and
                  Jonas Gehring and
                  Gabriel Synnaeve and
                  Taco Cohen},
  title        = {The KoLMogorov Test: Compression by Code Generation},
  booktitle    = {The Thirteenth International Conference on Learning Representations,
                  {ICLR} 2025, Singapore, April 24-28, 2025},
  publisher    = {OpenReview.net},
  year         = {2025},
  url          = {https://openreview.net/forum?id=C45YqeBDUM},
  bibsource    = {dblp computer science bibliography, https://dblp.org}
}

@article{grunwald2004,
  author       = {Peter Gr{\"{u}}nwald and
                  Paul M. B. Vit{\'{a}}nyi},
  title        = {Shannon Information and Kolmogorov Complexity},
  journal      = {CoRR},
  volume       = {cs.IT/0410002},
  year         = {2004},
  url          = {http://arxiv.org/abs/cs.IT/0410002},
  bibsource    = {dblp computer science bibliography, https://dblp.org}
}

@article{llmzip,
  author       = {Chandra Shekhara Kaushik Valmeekam and
                  Krishna Narayanan and
                  Dileep Kalathil and
                  Jean{-}Fran{\c{c}}ois Chamberland and
                  Srinivas Shakkottai},
  title        = {LLMZip: Lossless Text Compression using Large Language Models},
  journal      = {CoRR},
  volume       = {abs/2306.04050},
  year         = {2023},
  url          = {https://doi.org/10.48550/arXiv.2306.04050},
  doi          = {10.48550/ARXIV.2306.04050},
  eprinttype    = {arXiv},
  eprint       = {2306.04050},
  bibsource    = {dblp computer science bibliography, https://dblp.org}
}

@article{Depeiges2025,
  author       = {David Heurtel{-}Depeiges and
                  Anian Ruoss and
                  Joel Veness and
                  Tim Genewein},
  title        = {Compression via Pre-trained Transformers: {A} Study on Byte-Level
                  Multimodal Data},
  journal      = {CoRR},
  volume       = {abs/2410.05078},
  year         = {2024},
  url          = {https://doi.org/10.48550/arXiv.2410.05078},
  doi          = {10.48550/ARXIV.2410.05078},
  eprinttype    = {arXiv},
  eprint       = {2410.05078},
  bibsource    = {dblp computer science bibliography, https://dblp.org}
}

@inproceedings{schmidt2024,
  author       = {Craig W. Schmidt and
                  Varshini Reddy and
                  Haoran Zhang and
                  Alec Alameddine and
                  Omri Uzan and
                  Yuval Pinter and
                  Chris Tanner},
  editor       = {Yaser Al{-}Onaizan and
                  Mohit Bansal and
                  Yun{-}Nung Chen},
  title        = {Tokenization Is More Than Compression},
  booktitle    = {Proceedings of the 2024 Conference on Empirical Methods in Natural
                  Language Processing, {EMNLP} 2024, Miami, FL, USA, November 12-16,
                  2024},
  pages        = {678--702},
  publisher    = {Association for Computational Linguistics},
  year         = {2024},
  url          = {https://doi.org/10.18653/v1/2024.emnlp-main.40},
  doi          = {10.18653/V1/2024.EMNLP-MAIN.40},
  bibsource    = {dblp computer science bibliography, https://dblp.org}
}

@article{tay2021,
  author       = {Yi Tay and
                  Mostafa Dehghani and
                  Jinfeng Rao and
                  William Fedus and
                  Samira Abnar and
                  Hyung Won Chung and
                  Sharan Narang and
                  Dani Yogatama and
                  Ashish Vaswani and
                  Donald Metzler},
  title        = {Scale Efficiently: Insights from Pre-training and Fine-tuning Transformers},
  journal      = {CoRR},
  volume       = {abs/2109.10686},
  year         = {2021},
  url          = {https://arxiv.org/abs/2109.10686},
  eprinttype    = {arXiv},
  eprint       = {2109.10686},
  bibsource    = {dblp computer science bibliography, https://dblp.org}
}

@article{petty2023,
  author       = {Jackson Petty and
                  Sjoerd van Steenkiste and
                  Ishita Dasgupta and
                  Fei Sha and
                  Dan Garrette and
                  Tal Linzen},
  title        = {The Impact of Depth and Width on Transformer Language Model Generalization},
  journal      = {CoRR},
  volume       = {abs/2310.19956},
  year         = {2023},
  url          = {https://doi.org/10.48550/arXiv.2310.19956},
  doi          = {10.48550/ARXIV.2310.19956},
  eprinttype    = {arXiv},
  eprint       = {2310.19956},
  bibsource    = {dblp computer science bibliography, https://dblp.org}
}

@inproceedings{Inan2017,
  author       = {Hakan Inan and
                  Khashayar Khosravi and
                  Richard Socher},
  title        = {Tying Word Vectors and Word Classifiers: {A} Loss Framework for Language
                  Modeling},
  booktitle    = {5th International Conference on Learning Representations, {ICLR} 2017,
                  Toulon, France, April 24-26, 2017, Conference Track Proceedings},
  publisher    = {OpenReview.net},
  year         = {2017},
  url          = {https://openreview.net/forum?id=r1aPbsFle},
  bibsource    = {dblp computer science bibliography, https://dblp.org}
}

@misc{lesci2025,
      title={Causal Estimation of Tokenisation Bias}, 
      author={Pietro Lesci and Clara Meister and Thomas Hofmann and Andreas Vlachos and Tiago Pimentel},
      year={2025},
      eprint={2506.03149},
      archivePrefix={arXiv},
      primaryClass={cs.CL},
      url={https://arxiv.org/abs/2506.03149}, 
}

@inproceedings{Lee2022,
  author       = {Katherine Lee and
                  Daphne Ippolito and
                  Andrew Nystrom and
                  Chiyuan Zhang and
                  Douglas Eck and
                  Chris Callison{-}Burch and
                  Nicholas Carlini},
  editor       = {Smaranda Muresan and
                  Preslav Nakov and
                  Aline Villavicencio},
  title        = {Deduplicating Training Data Makes Language Models Better},
  booktitle    = {Proceedings of the 60th Annual Meeting of the Association for Computational
                  Linguistics (Volume 1: Long Papers), {ACL} 2022, Dublin, Ireland,
                  May 22-27, 2022},
  pages        = {8424--8445},
  publisher    = {Association for Computational Linguistics},
  year         = {2022},
  url          = {https://doi.org/10.18653/v1/2022.acl-long.577},
  doi          = {10.18653/V1/2022.ACL-LONG.577},
  bibsource    = {dblp computer science bibliography, https://dblp.org}
}

@inproceedings{Li2024,
  author       = {Jeffrey Li and
                  Alex Fang and
                  Georgios Smyrnis and
                  Maor Ivgi and
                  Matt Jordan and
                  Samir Yitzhak Gadre and
                  Hritik Bansal and
                  Etash Kumar Guha and
                  Sedrick Scott Keh and
                  Kushal Arora and
                  Saurabh Garg and
                  Rui Xin and
                  Niklas Muennighoff and
                  Reinhard Heckel and
                  Jean Mercat and
                  Mayee F. Chen and
                  Suchin Gururangan and
                  Mitchell Wortsman and
                  Alon Albalak and
                  Yonatan Bitton and
                  Marianna Nezhurina and
                  Amro Abbas and
                  Cheng{-}Yu Hsieh and
                  Dhruba Ghosh and
                  Josh Gardner and
                  Maciej Kilian and
                  Hanlin Zhang and
                  Rulin Shao and
                  Sarah M. Pratt and
                  Sunny Sanyal and
                  Gabriel Ilharco and
                  Giannis Daras and
                  Kalyani Marathe and
                  Aaron Gokaslan and
                  Jieyu Zhang and
                  Khyathi Raghavi Chandu and
                  Thao Nguyen and
                  Igor Vasiljevic and
                  Sham M. Kakade and
                  Shuran Song and
                  Sujay Sanghavi and
                  Fartash Faghri and
                  Sewoong Oh and
                  Luke Zettlemoyer and
                  Kyle Lo and
                  Alaaeldin El{-}Nouby and
                  Hadi Pouransari and
                  Alexander Toshev and
                  Stephanie Wang and
                  Dirk Groeneveld and
                  Luca Soldaini and
                  Pang Wei Koh and
                  Jenia Jitsev and
                  Thomas Kollar and
                  Alex Dimakis and
                  Yair Carmon and
                  Achal Dave and
                  Ludwig Schmidt and
                  Vaishaal Shankar},
  editor       = {Amir Globersons and
                  Lester Mackey and
                  Danielle Belgrave and
                  Angela Fan and
                  Ulrich Paquet and
                  Jakub M. Tomczak and
                  Cheng Zhang},
  title        = {DataComp-LM: In search of the next generation of training sets for
                  language models},
  booktitle    = {Advances in Neural Information Processing Systems 38: Annual Conference
                  on Neural Information Processing Systems 2024, NeurIPS 2024, Vancouver,
                  BC, Canada, December 10 - 15, 2024},
  year         = {2024},
  url          = {http://papers.nips.cc/paper\_files/paper/2024/hash/19e4ea30dded58259665db375885e412-Abstract-Datasets\_and\_Benchmarks\_Track.html},
  bibsource    = {dblp computer science bibliography, https://dblp.org}
}

\appendix
\appendix
\onecolumn

\section{Influence of token-frequency imbalance on unigram and language models} \label{apdx:motivation}

In this section, we provide a detailed explanation of our research question. Natural language exhibits strong contextual dependencies rather than behaving as an i.i.d. process: each token’s probability depends on its preceding context. 
As a result, the entropy rate, $H_\infty \;=\;\lim_{t\to\infty} H\bigl(X_t\mid X_{<t}\bigr)$, which captures the optimal per‐token uncertainty in text, is strictly lower than the unigram Shannon entropy
$H_1 \;=\; -\sum_{w} p(w)\,\log p(w)$. Although entropy rate and Shannon entropy coincide for truly i.i.d. data, in natural language, they can differ by several bits per token.

Under a pure unigram model, minimum cross‐entropy loss exactly equals Shannon entropy, and modifying the vocabulary size of the tokenizer immediately changes the Shannon entropy. Typically, enlarging the vocabulary segments frequent multi-token patterns into single tokens, driving their individual relative frequency up and reducing Shannon entropy. But if the vocabulary size grows too large, the new entries tend to be rare tokens, which lengthen the tail and can actually increase the Shannon entropy as token-frequency imbalance rises \cite{zouhar_tokenization}. However, experimental results show that expanding a BPE vocabulary to around $80\mathrm{K}$ lowers the Shannon entropy of unigram models, demonstrating that a more skewed token-frequency distribution is advantageous at practical vocabulary scales \cite{rajaraman2025theorytokenizationllms}. This pattern extends to n-gram models as well: their conditional Shannon entropy—the lowest possible n-gram cross-entropy—can never exceed the unigram entropy, so lowering the unigram entropy necessarily lowers the conditional entropy.

Language model loss $\mathcal L(\theta)$ minimizes

\begin{equation}\label{eq:loss}
\mathcal L(\theta)
=
H_{\infty}
\;+\;
\sum_{x\in V} p(x)\,
D_{\mathrm{KL}}\!\bigl(P(\cdot\mid x_{<t})\;\|\;Q_{\theta}(\cdot\mid x_{<t})\bigr).
\end{equation}

where \(V\) denotes the tokenizer’s vocabulary, \(p(x)\) the marginal probability of token \(x\), \(P(\,\cdot\mid x_{<t})\) the true next-token distribution given the full history \(x_{<t}\), and \(Q_\theta(\,\cdot\mid x_{<t})\) the model’s predicted distribution with parameters~\(\theta\). When the target label is a one-hot vector, the language model loss can be written as
$
\mathcal{L}(\theta)
=\sum_{x\in V}p(x)\,\bigl[-\log Q_\theta(x\mid x_{<t})\bigr]$ so it is a marginal‐frequency‐weighted average of the model’s surprisal \(-\log Q_\theta\).  By contrast, the Shannon entropy of the unigram model is a marginal‐frequency‐weighted average of the self-information \(-\log p(x)\) in the dataset. Even though frequent token logits and embedding norms are higher than rare ones, the loss \(-\log Q_\theta\) depends not only on the underlying relative token-frequencies but also on training dynamics as well. We therefore cannot derive a closed-form expression to predict how much the loss on frequent tokens shrinks versus how much the rare token losses grow under $\mathcal{L}= p_{\mathrm{freq}}\,L_{\mathrm{freq}} + p_{\mathrm{rare}}\,L_{\mathrm{rare}}$. This masking effect makes it substantially harder to measure the influence of token‐frequency imbalance in language models than in the unigram and n-gram models.


\newpage

\section{Coverage of the most frequent N words}\label{apdx:coverage}

\begin{figure}[htbp]
  \centering
  \begin{subfigure}[b]{0.48\textwidth}
    \centering
    \includegraphics[width=\textwidth]{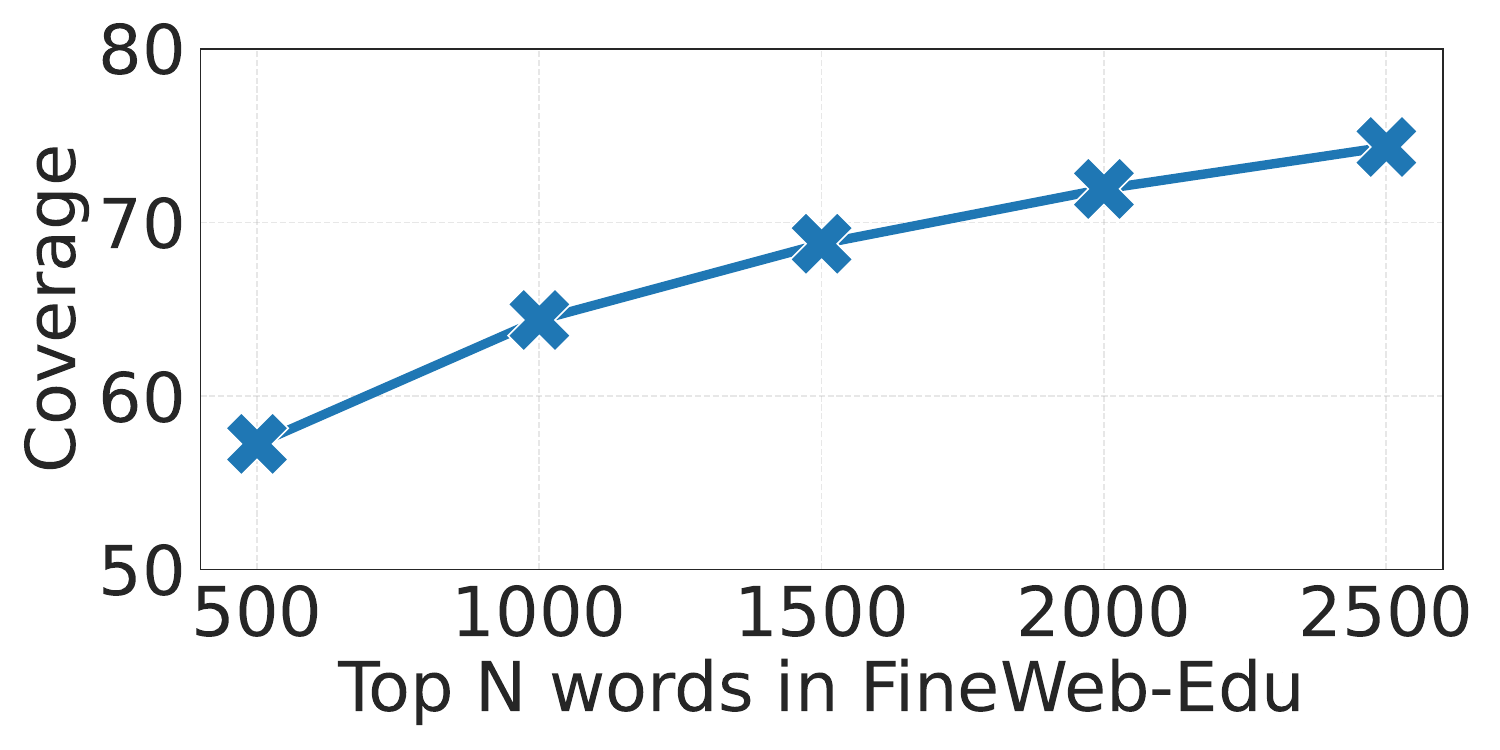}
    \caption{Coverage of frequent words in FineWeb-Edu}
    \label{fig:6a}
  \end{subfigure}
  \hfill
  \begin{subfigure}[b]{0.48\textwidth}
    \centering
    \includegraphics[width=\textwidth]{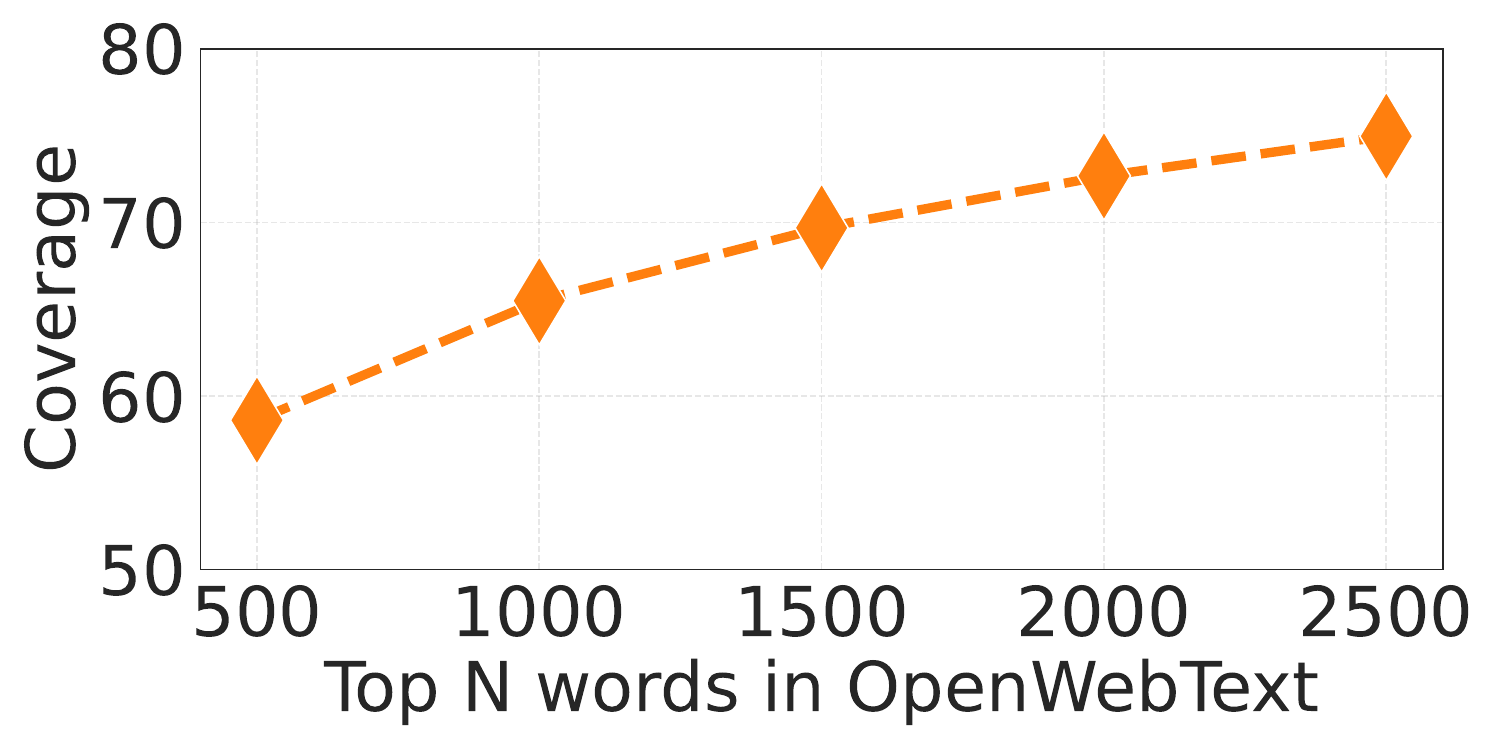}
    \caption{Coverage of frequent words in OpenWebText}
    \label{fig:6b}
  \end{subfigure}

  \caption{Figures \ref{fig:6a} and \ref{fig:6b} illustrate the cumulative coverage of the 2,500 most frequent words in the Fineweb-edu and OpenWebText datasets, respectively.}
  \label{fig:coverage}
\end{figure}

In this section, we measure the coverage of the frequent words in Fineweb-edu \cite{penedo2024finewebdatasetsdecantingweb} and OpenWebText \cite{Gokaslan2019OpenWeb}. Both dataset exhibit a steep rise in cumulative coverage as we include more high-frequency tokens, but with subtly different baselines and slopes. In figure \ref{fig:6a}, the most frequent $500$ words already cover about $58$\% of all tokens, climbing to roughly $75$\% once we take the $2,500$ most frequent words. OpenWebText (Figure \ref{fig:6b}) starts marginally higher—around $59$\% at most frequent $500$ words, but follows an almost identical trajectory, reaching about $76$\% coverage by the frequent $2,500$ words. This pattern underscores how a relatively small core vocabulary captures the vast majority of running text in both corpora, with only modest gains as we move deeper into the long tail.

\section{OpenWebText experiments results}\label{apdx:openwebtext}

\begin{figure}[htbp]
  \centering

  \begin{subfigure}[b]{0.48\textwidth}
    \centering
    \includegraphics[width=\textwidth]{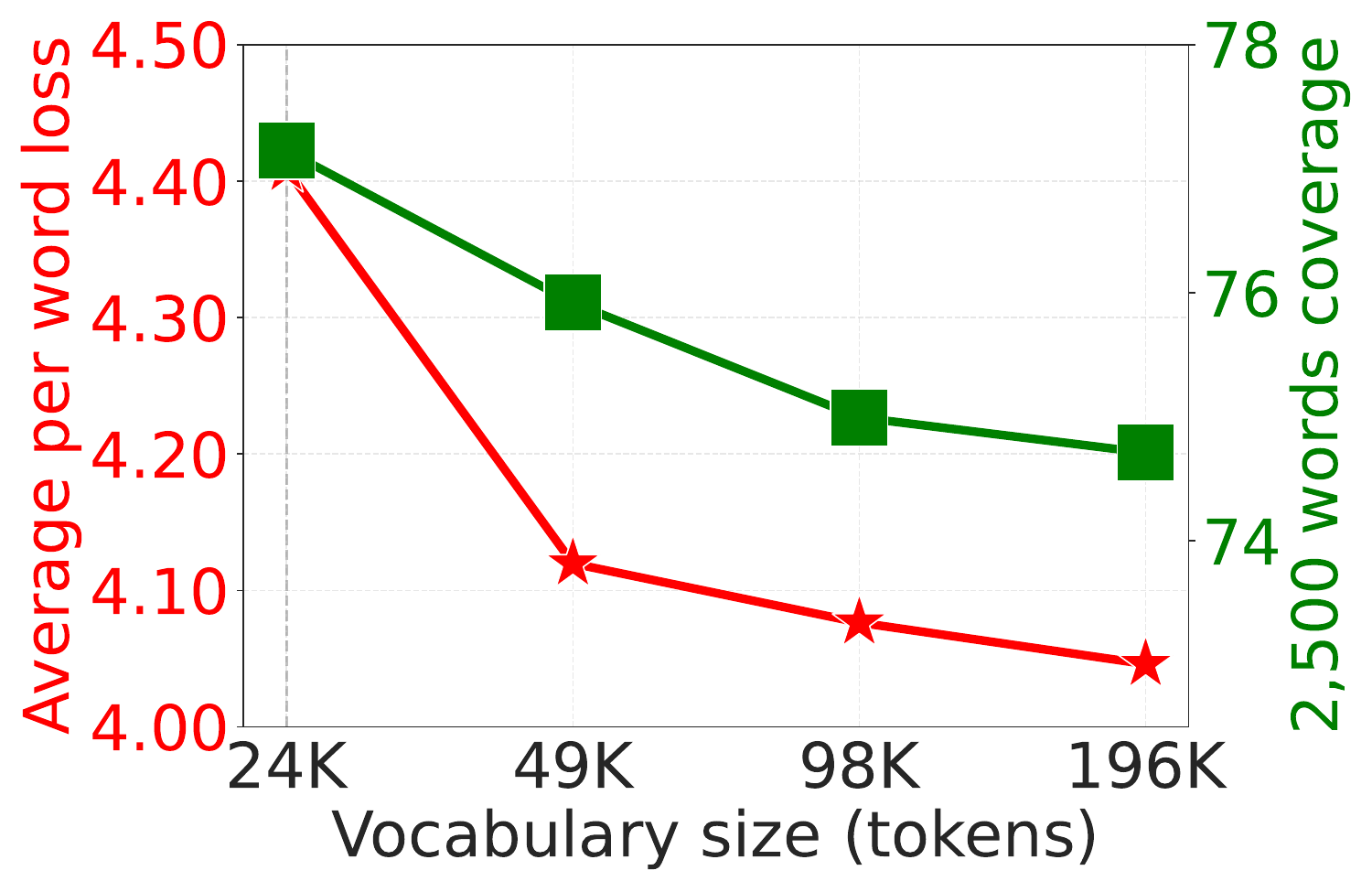}
    \caption{Frequent word loss on an OpenWebText}
    \label{fig:7a}
  \end{subfigure}
  \hfill
  \begin{subfigure}[b]{0.48\textwidth}
    \centering
    \includegraphics[width=\textwidth]{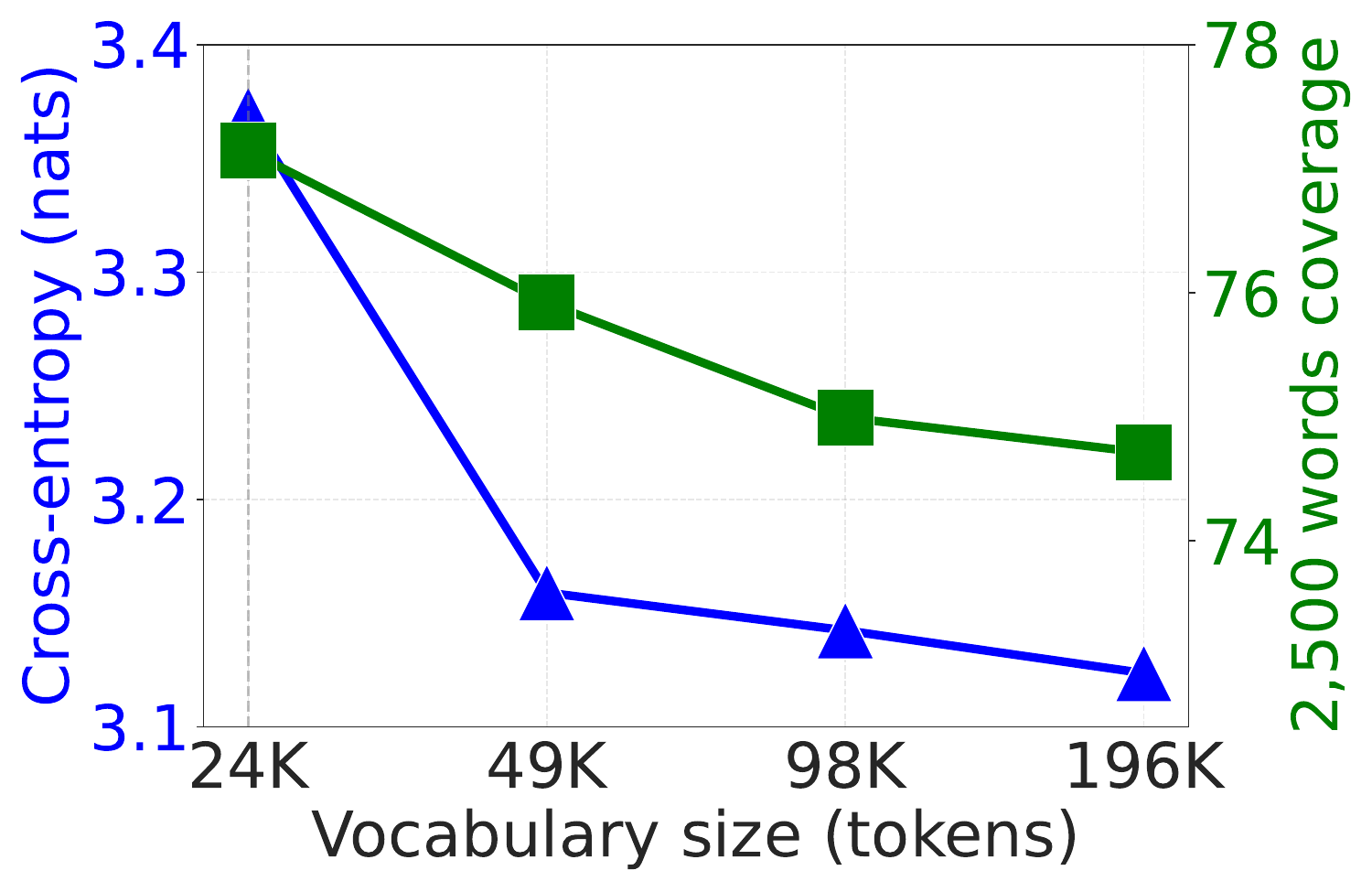}
    \caption{Global cross-entropy on an OpenWebText}
    \label{fig:7b}
  \end{subfigure}

  \caption{Figure \ref{fig:7a} reveals that expanding the vocabulary from $24\mathrm{K}$ to $196\mathrm{K}$ steadily reduces the average per-vocabulary loss of high frequency words. Figure \ref{fig:7b} indicates that the most frequent $2,500$ tokens still contribute roughly $75$\ of the total loss, while the rare words losses grow with vocabulary size, similar to \ref{fig:3b}. Figure \ref{fig:7b} further shows that the global cross-entropy loss falls by about $0.25$ nats as the vocabulary grows, demonstrating that the reduction of loss on frequent words outweighs the inflation of rare-token losses.}
  \label{fig:web}
\end{figure} 

To verify that reducing frequent-word loss is not a by-product of dataset quality, we repeat the same experiments in section \ref{main:token freq} on the OpenWebText dataset. Figure \ref{fig:7a} shows that widening the vocabulary from $24\mathrm{K}$ to $196\mathrm{K}$ in OpenWebText progressively reduces the average loss assigned to high-frequency words. Figure \ref{fig:7b} indicates that the most frequent $2,500$ tokens still account for about $75$\ of the total loss, whereas the loss on infrequent tokens grows with vocabulary size, paralleling the pattern seen in Figure \ref{fig:3b}. Figure \ref{fig:7b} further demonstrates that the global cross-entropy loss falls from $3.37$ nats at a $24\mathrm{K}$ vocabulary to $3.12$ nats at $196\mathrm{K}$, implying that the reduction in loss on frequent words more than offsets the increase in rare-token loss, regardless of dataset quality or type.

\newpage

\section{OLMo-2 result} \label{apdx:olmo}

\begin{figure}[htbp]
  \centering

  \begin{subfigure}[b]{0.48\textwidth}
    \centering
    \includegraphics[width=\textwidth]{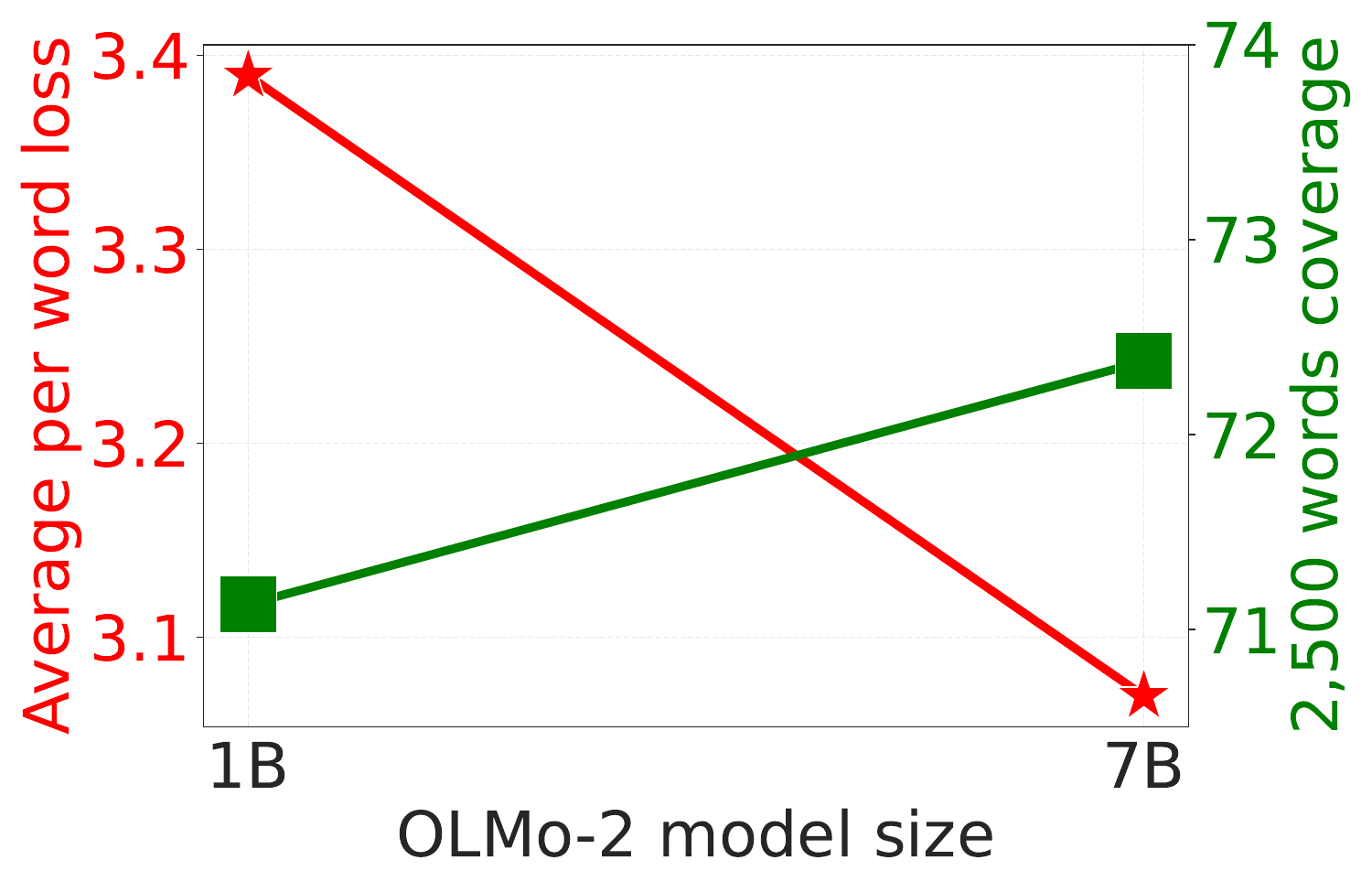}
    \caption{Frequent word loss in OLMo-2 models}
    \label{fig:8a}
  \end{subfigure}
  \hfill
  \begin{subfigure}[b]{0.48\textwidth}
    \centering
    \includegraphics[width=\textwidth]{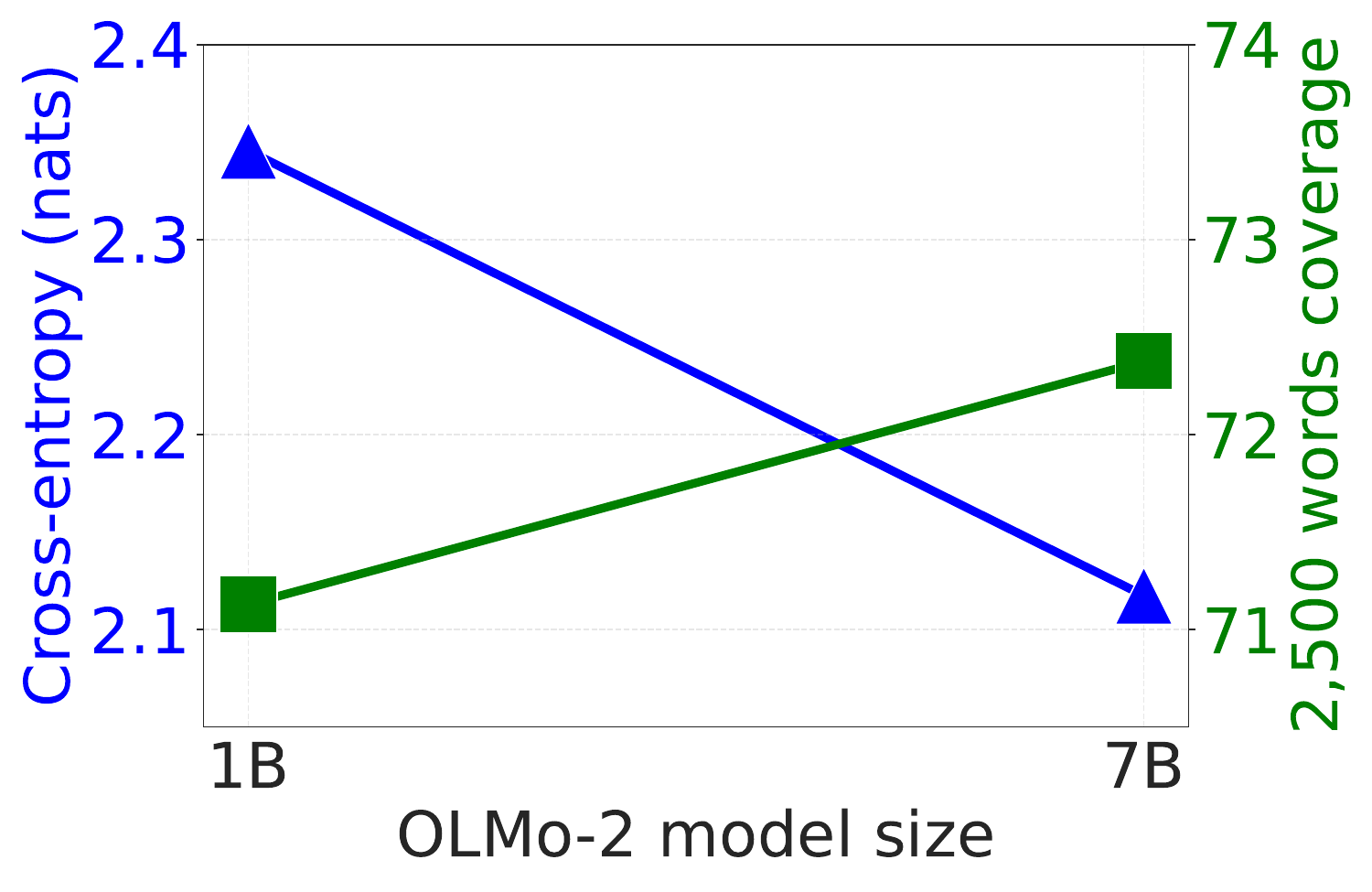}
    \caption{Global cross-entropy in OLMo-2 models}
    \label{fig:8b}
  \end{subfigure}
  
  \caption{Figure \ref{fig:8a} illustrates that larger $7\mathrm{B}$ model reduces average per-vocabulary loss from $3.39$ nats to $3.07$ nats while slightly increasing the proportion of loss covered by frequent words from $71\%$ to $72.5\%$. Figure \ref{fig:8b} further demonstrates that Scaling from $1\mathrm{B}$ to $7\mathrm{B}$ reduces the overall cross-entropy from $2.35$ nats to $2.12$ nats, confirming the same pattern persists in contemporary large-scale language models.}
  \label{fig:olmo size}
\end{figure}

To identify whether the reduction in frequent words loss with increasing model size holds for contemporary large language models, we perform analogous experiments using the OLMo-2 series \cite{olmo2}. Figure \ref{fig:8a} and \ref{fig:8b} indicate that the average per vocabulary loss falls from $3.39$ nats in the $1$B parameter model to $3.07$ nats in the $7$B variant while slightly increasing the proportion of loss covered by $2,500$ frequent words from $71\%$ to $72.5\%$. Figure \ref{fig:8b} further shows that global cross-entropy loss declines from $2.35$ nats for OLMo-2 $1\mathrm{B}$ to $2.12$ nats for OLMo-2 $7$B. Notably, OLMo-2 employs a much larger vocabulary (cl100K \cite{gpt-4}) than Pythia (50304 tokens \cite{gpt-neox}) and trains on a larger corpus \cite{olmo2}, which helps drive down the average loss on high-frequency words. These results confirm that the same trend holds for modern large-scale language models as well.

\section{Are the gains mainly driven by parameter increases from vocab expansion?}

To probe whether gains come from parameter growth with larger vocabularies, we run a controlled study comparing a $24\mathrm{K}$ vocabulary model ($122\mathrm{M}$ parameters) to a near equal size $49\mathrm{K}$ model ($124\mathrm{M}$ parameters) to isolate embedding capacity effects. We reduced the hidden dimension from $768$ to $648$ and the feed‑forward dimension from $2048$ to $1728$, while maintaining identical model depth which known to have a dominant impact on performance \cite{tay2021, petty2023}. As shown in the table \ref{tab:parametergrowth}, the $49\mathrm{K}$ model incurs higher cross‑entropy loss and delivers inferior results on downstream benchmarks (ARC‑E, HellaSWAG, PIQA, and SCIQ) compared to its $24\mathrm{K}$ counterpart, indicating that simply expanding the embedding layer contributes far less to model expressivity than is often assumed.

\begin{table}[H]
  \centering
  \caption{A $24\mathrm{K}$ vs. size-matched $49\mathrm{K}$ indicates embedding expansion yields limited gains.}
  \label{tab:parametergrowth}
  \begin{tabular}{lcc}
    \toprule
    \textbf{Vocab (Params)} & \textbf{CE Loss} & \textbf{Downstream (\%)} \\
    \midrule
    24K (122M) & 3.179 & 54.73 \\
    49K (124M) & 3.563 & 50.68 \\
    49K (161M) & 3.171 & 55.44 \\
    \bottomrule
  \end{tabular}
\end{table}

\newpage

\section{Do the benefits of larger vocabularies persist with tied-embedding models?}

Several studies show that tying input and output embeddings can change pre-training dynamics \cite{Inan2017,lesci2025}. We therefore repeat our study with tied embeddings and report average frequent-word loss and cross-entropy on FineWeb-Edu in the table \ref{tab:tiedembd}. The tied embedding variant shows broadly similar average frequent word loss and ce-loss to the untied baseline.

\begin{table}[H]
  \centering
  \caption{Average frequent-word loss and cross-entropy of tied and untied embedding models}
  \label{tab:tiedembd}
  \begin{tabular}{llrrrr}
    \toprule
    \multicolumn{2}{c}{} & \textbf{24K} & \textbf{49K} & \textbf{98K} & \textbf{196K} \\
    \midrule
    \multirow{2}{*}{\textbf{Tied}}
      & Avg frequent word loss & 4.067 & 4.029 & 3.991 & 3.968 \\
      & Cross entropy          & 3.177 & 3.160 & 3.146 & 3.134 \\
    \midrule
    \multirow{2}{*}{\textbf{Untied}}
      & Avg frequent word loss & 4.074 & 4.031 & 4.001 & 3.977 \\
      & Cross entropy          & 3.179 & 3.163 & 3.147 & 3.136 \\
    \bottomrule
  \end{tabular}
\end{table}

\section{To what extent does a larger model decrease frequent and rare word loss?}

We conducted experiments to quantify scaling effects on frequent and rare words loss and demonstrate that scaling model capacity disproportionately benefits the prediction of high‑frequency tokens relative to low‑frequency tokens. Since BPE vocabulary IDs are not token frequency ordered, we cannot directly distinguish common and rare tokens in the Pythia and OLMo‑2 training corpora. To proxy frequency, we utilize the Google Web Trillion Word Corpus frequency list, which records counts for $333,333$ unique words appearing at least $10,000$ times in a $1.0249\times10^{12}$ words. We classify the top $2,500$ entries as frequent words and the bottom $10,000$ as rare words. As shown in the table \ref{tab:freq-rare-model}, although cross‑entropy loss declines for both groups with increasing model size, the reduction for frequent tokens is markedly greater. We therefore conclude that the dominant effect of model scaling is the lowering of frequent‑token loss, which in turn drives the overall cross‑entropy reduction.

\begin{table}[H]
  \centering
  \caption{Frequent, rare word loss and cross-entropy loss of Pythia and OLMo-2}
  \label{tab:freq-rare-model}
  \begin{tabular}{l l c c c}
    \toprule
    \textbf{Model} & \textbf{Params} & \textbf{Frequent word loss} & \textbf{Rare word loss} & \textbf{CE-loss} \\
    \midrule
    \multirow{2}{*}{Pythia} & 1B & 3.01 & 4.99 & 2.51 \\
                            & 7B & 2.55 & 4.93 & 2.26 \\
    \addlinespace[0.25em]
    \multirow{2}{*}{OLMo-2} & 1B & 2.60 & 4.27 & 2.22 \\
                            & 7B & 2.20 & 4.19 & 2.00 \\
    \bottomrule
  \end{tabular}
\end{table}

\newpage

\section{Learning rate exploration}
We use a fixed learning rate of $6\times10^{-4}$
for an $85\mathrm{M}$ non-embedding–parameter language model, mirroring the learning rate of GPT-3 with an equivalent non-embedding model size \cite{gpt-3}. we further conducted a learning‑rate exploration with four additional learning rates: $2.4\times10^{-3}$, $1.2\times10^{-3}$, $1.5\times10^{-4}$, $7.5\times10^{-5}$. Table \ref{tab:ce_loss_lrs} and \ref{tab:freq_loss_lrs} report the resulting cross‑entropy loss and average word loss of frequent 2500 words in fineweb-edu across different vocabulary sizes and learning rates, using the same validation dataset described in Section \ref{main:setting}. 

\begin{table}[H]
  \centering
  \caption{Cross-entropy loss across five learning rates.}
  \label{tab:ce_loss_lrs}
  \begin{tabular}{lccccc}
    \toprule
    \textbf{Cross-Entropy Loss} & \textbf{2.4e-3} & \textbf{1.2e-3} & \textbf{6.0e-4} & \textbf{1.5e-4} & \textbf{7.5e-5} \\
    \midrule
    24K  & 3.156 & 3.141 & 3.179 & 3.484 & 3.779 \\
    49K  & 3.135 & 3.114 & 3.163 & 3.461 & 3.751 \\
    98K  & 3.122 & 3.099 & 3.147 & 3.443 & 3.736 \\
    196K & 3.111 & 3.086 & 3.136 & 3.429 & 3.725 \\
    \bottomrule
  \end{tabular}
\end{table}

\begin{table}[H]
  \centering
  \caption{Average frequent-word loss across five learning rates.}
  \label{tab:freq_loss_lrs}
  \begin{tabular}{lccccc}
    \toprule
    \textbf{Frequent Word Loss} & \textbf{2.4e-3} & \textbf{1.2e-3} & \textbf{6.0e-4} & \textbf{1.5e-4} & \textbf{7.5e-5} \\
    \midrule
    24K  & 4.033 & 4.012 & 4.074 & 4.431 & 4.783 \\
    49K  & 3.990 & 3.970 & 4.031 & 4.399 & 4.749 \\
    98K  & 3.963 & 3.939 & 4.001 & 4.373 & 4.725 \\
    196K & 3.942 & 3.916 & 3.977 & 4.352 & 4.703 \\
    \bottomrule
  \end{tabular}
\end{table}

\section{Detailed Experimental Setting} \label{apdx:setting}

In this section, we provide detailed configurations of pretraining to reproduce our results. The training setup (Table~\ref{tab:train-config}) uses a global batch size of 256, weight decay 0.1, and sequence length 2048. Optimization is Adam with a cosine learning-rate schedule, a \(700\)-step warmup, and a weight-initialization scale of 0.02. The model setup (Table~\ref{tab:model-config}) covers two different model size: an 85M model with 12 layers and 12 heads \((d_{\text{model}}=768,\; d_{\text{ffn}}=2048,\; d_{\text{head}}=64)\) and a 450M model with 21 layers and 21 heads \((d_{\text{model}}=1344,\; d_{\text{ffn}}=3548,\; d_{\text{head}}=21)\). Together, these tables specify the standardized training hyperparameters and the core architectural dimensions for both scales.

\begin{table}[H]
  \centering
  \caption{Training configurations. LR Schedule denotes learning-rate schedule.}
  \label{tab:train-config}

  \begin{tabular}{lccc}
    \toprule
    \textbf{Global Batch Size} & \textbf{Weight Decay} & \textbf{Sequence Length} & \textbf{Optimizer} \\
    \midrule
    256 & 0.1 & 2048 & AdamW \\
    \bottomrule
  \end{tabular}

  \vspace{0.4em}

  \begin{tabular}{lcc}
    \toprule
    \textbf{LR Schedule} & \textbf{Warmup} & \textbf{Weight Init.} \\
    \midrule
    Cosine & 700 steps & 0.02 \\
    \bottomrule
  \end{tabular}
\end{table}

\begin{table}[H]
  \centering
  \caption{Model configurations.}
  \label{tab:model-config}
  \begin{tabular}{lccccc}
    \toprule
    \textbf{Size} & \(n_{\text{layers}}\) & \(n_{\text{heads}}\) & \(d_{\text{model}}\) & \(d_{\text{ffn}}\) & \(d_{\text{head}}\) \\
    \midrule
    85M & 12 & 12  & 768 & 2048 & 64 \\
    450M & 21 & 21 & 1344 & 3548 & 21 \\

    \bottomrule
  \end{tabular}
\end{table}

\end{document}